\documentclass[journal]{IEEEtran}
\usepackage{amsmath,amsfonts}
\usepackage{lineno} % To obtain line numbers
\usepackage{algorithm}
\usepackage{algpseudocode}
\usepackage{array}
\usepackage[caption=false,font=normalsize,labelfont=sf,textfont=sf]{subfig}
\usepackage{textcomp}
\usepackage{stfloats}
\usepackage{url}
\usepackage{verbatim}
\usepackage{graphicx}

\usepackage{xcolor}

\usepackage{amssymb}
\usepackage{xcolor}
\usepackage{flushend}

\usepackage{booktabs,makecell,tabularx}
\usepackage{multirow}
\usepackage[export]{adjustbox} %centering figure
\usepackage{orcidlink}
\usepackage{footmisc}

\usepackage{epstopdf}
\usepackage{epsfig}

\usepackage{cite}

\usepackage{textcase}
\usepackage{xcolor}
\usepackage{float}

\newcommand{\ModelName}{R-Sparse R-CNN}
\newcommand{\SparseRCNN}{Sparse R-CNN}

\newcommand{\InteractionHead}{\textit{Interaction Head}}
\newcommand{\InteractionHeads}{\textit{Interaction Heads}}
\newcommand{\BackgroundHead}{\textit{Background Interaction Head}}
\newcommand{\ObjectHead}{\textit{Object Interaction Head}}
\newcommand{\FusionHead}{\textit{Fusion Head}}
\newcommand{\InteractionModule}{\textit{Interaction Module}}

\newcommand{\DetectionHead}{\textit{Detection Head}}
\newcommand{\DetectionHeads}{\textit{Detection Heads}}

\newcommand{\DCP}{\textit{Dual-Context Pooling}}
\newcommand{\ProposalType}{background-aware proposals}
\newcommand{\ProposalTypeAbbr}{BAPs}
\newcommand{\DynInstInteraction}{\textit{Dynamic Instance Interaction}}

\newcommand{\Figure}{Fig.}
\newcommand{\Equation}{Eq.}
\newcommand{\Table}{Table}

\newcommand{\Section}{Section}
\newcommand{\EtAl}{\textit{et al.}}

\newcommand{\colA}[1]{\textcolor{black}{#1}} % color for Reviewer#1
\newcommand{\colB}[1]{\textcolor{black}{#1}} % color for Reviewer#2
\newcommand{\colC}[1]{\textcolor{black}{#1}} % color for Reviewer#3
\newcommand{\colD}[1]{\textcolor{black}{#1}} % color for Reviewer#4

\begin{document}

\title{\ModelName: SAR Ship Detection Based on Background-Aware Sparse Learnable Proposals}

\author{
Kamirul Kamirul~\orcidlink{0000-0002-1474-1139},~\IEEEmembership{Student Member,~IEEE},
Odysseas A. Pappas~\orcidlink{0000-0003-3037-2828}, and
Alin M. Achim~\orcidlink{0000-0002-0982-7798},~\IEEEmembership{Senior Member,~IEEE}

\thanks{
    Kamirul Kamirul, Odysseas A. Pappas, and Alin M. Achim are with the
    Visual Information Laboratory, University of Bristol, BS1 5DD Bristol,
    U.K. (e-mail: kamirul.kamirul@bristol.ac.uk; o.pappas@bristol.ac.uk;
    alin.achim@bristol.ac.uk). (Corresponding author: Kamirul Kamirul.)
}
\thanks{*The code will be made public upon the publication of the paper.}
\thanks{Manuscript received February xx, 2025; revised February xx, 2025.}
}

\markboth{}%
{Shell \MakeLowercase{\textit{et al.}}: A Sample Article Using IEEEtran.cls for IEEE Journals}


\maketitle

\begin{abstract}
We introduce~\ModelName, a novel pipeline for oriented ship detection in \colA{Synthetic Aperture Radar (SAR)} images that leverages sparse learnable proposals enriched with background contextual information, termed~\ProposalType~(\ProposalTypeAbbr).
The adoption of sparse proposals streamlines the pipeline by eliminating the need for proposal generators and post-processing for overlapping predictions.
\colA{
The proposed~\ProposalTypeAbbr~enrich object representation by integrating ship and background features, allowing the model to learn their contextual relationships for more accurate distinction of ships in complex environments.}
To complement~\ProposalTypeAbbr, we propose\colA{~\DCP~(DCP)}, a novel strategy that jointly extracts ship and background features in a single unified operation.
This unified design improves efficiency by eliminating redundant computation inherent in separate pooling.
Moreover, by ensuring that ship and background features are pooled from the same feature map level, DCP provides aligned features that improve contextual relationship learning.
\colA{
Finally, as a core component of contextual relationship learning in~\ModelName, we design a dedicated transformer-based~\InteractionModule.
This module interacts pooled ship and background features with corresponding proposal features and models their relationships.}
\colC{Experimental} results show that~\ModelName~delivers outstanding accuracy, surpassing state-of-the-art models by margins of up to 12.8\% and 11.9\% on SSDD and RSDD-SAR inshore datasets, respectively. 
These results demonstrate the effectiveness and competitiveness of~\ModelName~as a robust framework for oriented ship detection in SAR imagery.
The code is available at: \textcolor{blue}{\url{www.github.com/ka-mirul/R-Sparse-R-CNN}}*. 
\end{abstract}

\begin{IEEEkeywords}
Background-aware proposals, convolutional neural networks, deep learning,
oriented ship detection, sparse learnable proposals, synthetic
aperture radar.
\end{IEEEkeywords}

\section{Introduction}
\IEEEPARstart{S}{ynthetic} Aperture Radar (SAR) is a powerful active microwave imaging technology that functions reliably under all weather conditions and at any time of day. 
Among its various applications, the ship detection task is one of the most prevalent, applicable for both civilian and military maritime monitoring purposes.

\colA{
SAR ship detection has traditionally relied on rule-based techniques that use hand-crafted features and statistical modeling.
Methods such as Constant False Alarm Rate (CFAR) detection have been widely employed to distinguish ships from sea clutter by applying adaptive thresholding~\cite{Multilayer_CFAR, Rayleigh_Mixture_CFAR, Censored_Harmonic_CFAR, BTS_CFAR, OS_CFAR, Bilateral_CFAR, TS_CFAR, SP-CFAR_Odysseas, SPCFAR_Target_Detection_SAR, Improved_SPCFAR_HighRes_Target, SPCFAR_Truncated_Gamma}.
Various adaptations of CFAR techniques have been designed to match specific conditions of sea background interference, including Multilayer CFAR~\cite{Multilayer_CFAR}, RmSAT-CFAR~\cite{Rayleigh_Mixture_CFAR}, CHA-CFAR~\cite{Censored_Harmonic_CFAR}, BTS-RCFAR~\cite{BTS_CFAR}, OS-CFAR~\cite{OS_CFAR}, Bilateral CFAR~\cite{Bilateral_CFAR}, TS-CFAR~\cite{TS_CFAR}, and several superpixel-based CFAR methods~\cite{SP-CFAR_Odysseas, SPCFAR_Target_Detection_SAR, Improved_SPCFAR_HighRes_Target, SPCFAR_Truncated_Gamma}.
Beyond conventional approaches that rely on ship signatures, recent approaches have explored wake signatures as indirect cues for detecting ship presence~\cite{Murphy_Radon_051986, Rey_RAdon_07_1990, COURMONTAGNE_Radon_04_2025, Karakus_Ship_Wake_052019}.
The development of simulated SAR wakes~\cite{Tunaley_1991, Nunziata, Facet_Approach_032012, RIZAEV2022120, Kamirul_SAR_Ship_S_Band, SynthwakeSAR_Paper} further advances this direction by offering theoretically infinite datasets. 
Despite their effectiveness, detecting ships and their wake signatures in SAR images are still heavily reliant on hand-crafted features, which limits their adaptability to varying imaging conditions.
CFAR-based techniques, for example, perform well in homogeneous sea clutter but struggle in more complex coastal regions. 
Additionally, the fixed protective window sizes in CFAR methods hinder their ability to generalize across ships of different sizes, often leading to missed detections, especially in dense maritime traffic scenarios.
These limitations have sparked a shift towards Convolutional Neural Network (CNN)-based approaches.
}

% In traditional SAR ship detection systems, methods based on the Constant False Alarm Rate (CFAR) technique are commonly employed. 
% CFAR detectors adaptively estimate sea clutter statistics around a target based on an assumed background probability distribution, ensuring a consistent probability of false alarm (PFA) \cite{Multilayer_CFAR, Rayleigh_Mixture_CFAR, Censored_Harmonic_CFAR, BTS_CFAR, OS_CFAR, Bilateral_CFAR, TS_CFAR, SP-CFAR_Odysseas, SPCFAR_Target_Detection_SAR, Improved_SPCFAR_HighRes_Target, SPCFAR_Truncated_Gamma}.

\begin{figure}[ht]
\includegraphics[width=0.49\textwidth,center, trim={0 0.3cm 0 0}]{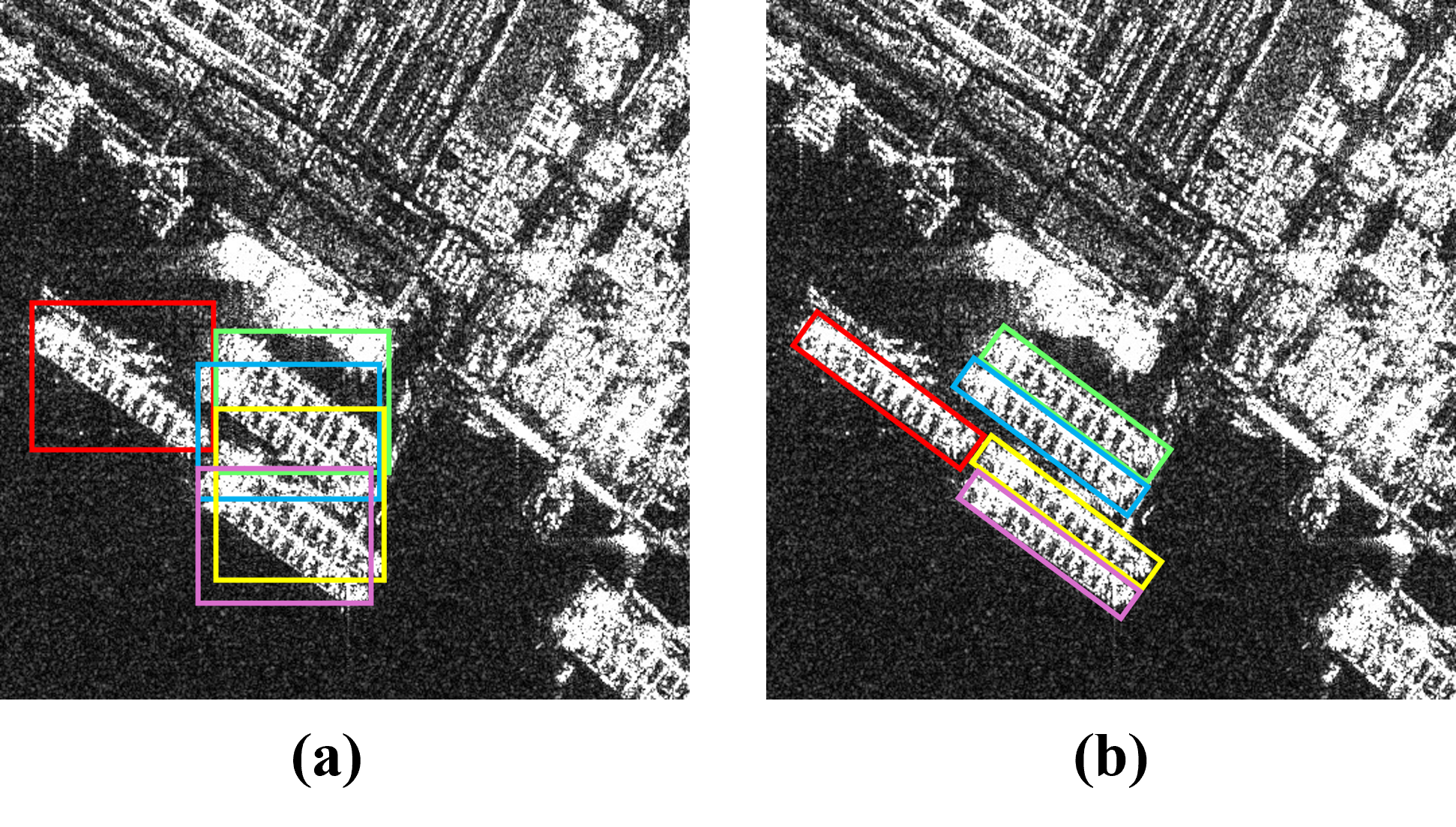}
\caption{Illustration of ships detected by (a) HBB and (b) OBB detectors in an inshore SAR image scene.}
\label{fig:HBB_vs_OBB}
\end{figure}

% With the advancement of CNN, there is a strong trend toward their adoption in ship detection tasks in SAR imagery. 
Most CNN-based SAR ship detectors draw inspiration from common object detection techniques, i.e., \colA{horizontal bounding box (HBB)} detection~\cite{YOLO-v1, SSD_paper, Faster-RCNN-original-paper, CornerNet, CenterNet, FCOS}.
While HBB detectors provide accurate localization, their application for detecting oriented ships is inherently suboptimal.
HBBs often encompass not only the ships but also significant portions of their surroundings.
This introduces redundant information and excessive background interference. 
\colA{
The limitation of HBBs becomes even more pronounced in scenarios such as densely docked ships in inshore regions (as illustrated in ~\Figure{\ref{fig:HBB_vs_OBB}a}), where overlapping bounding boxes are frequently produced.
As a result, a single box may enclose multiple ships or land features, potentially reducing detection accuracy.
}

\colA{
To address the abovementioned limitations of horizontal bounding box (HBB) detectors, recent studies have increasingly focused on Oriented Bounding Box (OBB) detectors~\cite{Guo_Ship_OBB_2023, YOLOv5_BiFPN, YOLO-Lite_Lighweight, YOLOv3-Tiny_RealTime, Ma_GF3, CoAM_RFIM, DCMSNN_2018, ARPN_2020, Faster_R_CNN_Ship_2021, Cascade_RCNN_ship_2024, Mask_RCNN_MultiTemporal_102019, SiS_SARShip_MaskRCNN_012023, FBR-Net_072020, CenterNet++_Ship_042021, Ship_FCOS_HRS_072021, FCOS_Ship_Large-Scale_022022}.
Unlike HBBs, OBBs provide a more precise encapsulation of ship features while effectively capturing ship orientation as illustrated in~\Figure{\ref{fig:HBB_vs_OBB}b}.
This orientation information is crucial for understanding vessel direction, enabling advancements in route prediction, collision avoidance systems, and maritime traffic analysis.
}

\colA{In CNN-based object detection, anchor-based methods have established themselves as a dominant framework owing to their ability to effectively handle objects with varying shapes, sizes, and orientations~\cite{SAR-Ship_Improved_RCNN_2017, DCMSNN_2018, Squeeze_Excitation_Network_2018, ARPN_2020, Lighweight_Faster_RCNN_2022}. 
% These methods densely generate anchor boxes over feature maps, serving as initial guesses for object bounding across various scales, aspect ratios, and orientations. 
% A network then regresses these anchors to align with ground truth bounding boxes while simultaneously predicting class probabilities.
The predefined nature of anchor boxes enables precise object localization and makes anchor-based detectors robust in complex detection scenarios. 
In contrast, anchor-free methods, which directly predict bounding boxes without relying on anchor boxes, often struggle to achieve comparable accuracy and reliability.} 

Despite their advantages, anchor-based approaches suffer from two notable drawbacks: (1) overlapping predictions and (2) sensitivity to anchor configurations.
The dense generation of anchors leads to substantial overlap among predicted boxes, requiring \colA{non-maximum suppression (NMS)} to resolve.
This increases computational burden and slows inference~\cite{SoftNMS}. 
Furthermore, the detection performance of anchor-based detectors is highly dependent on initial anchor properties such as the number, size, and aspect ratios~\cite{Focal_Loss, YOLO9000, Faster-RCNN-original-paper}. 
To mitigate these issues, recent research has focused on reducing NMS reliance and minimizing the model's dependency on dense anchors, giving rise to sparse proposal methods~\cite{GCNN, DETR, Sparse-RCNN, Sparse_Anchoring}.

Sparse proposal-based detectors simplify the process by eliminating the need to design dense anchors as proposal candidates.
G-CNN~\cite{GCNN} can be considered a forerunner to this class of detectors. Instead of utilizing a proposal generator, this approach initiates with a multi-scale grid of 180 fixed bounding boxes. 
A regressor is then trained to iteratively refine the position and scale of these grid elements to better align with the objects.
At that time, G-CNN delivered performance on par with Fast R-CNN~\cite{Fast-RCNN-original-paper}, which uses around 2,000 bounding boxes.
However, G-CNN trails in performance compared to the current successor of Fast R-CNN, Faster R-CNN~\cite{Faster-RCNN-original-paper}.

\begin{figure*}[t]
\includegraphics[width=0.99\textwidth,center, trim={0 0 0.9 0}]{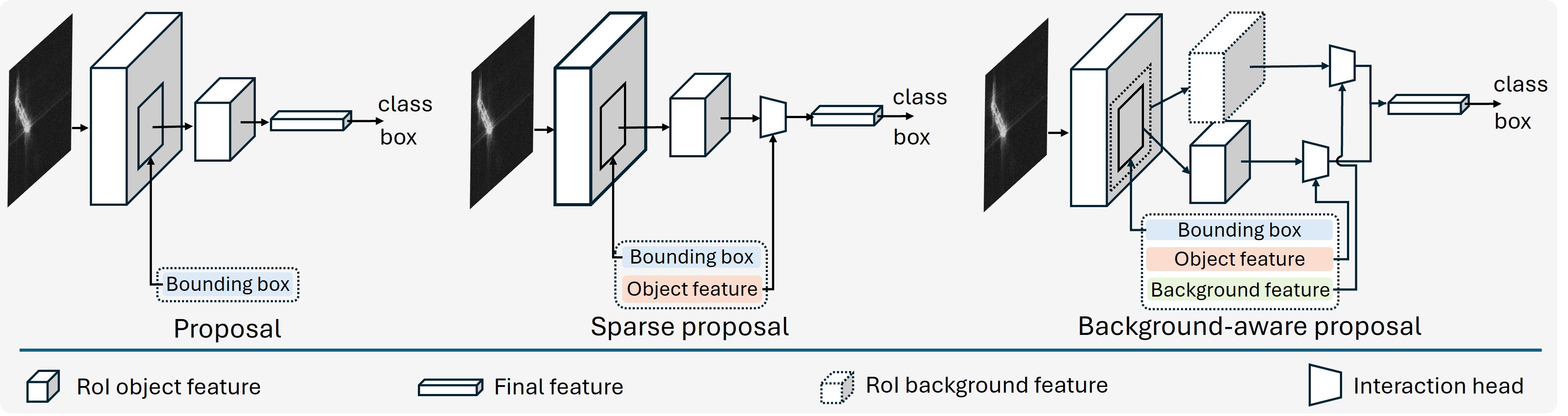}
\caption{
    Comparison of proposals on object detection pipelines: (left) traditional proposals in Faster R-CNN~\cite{Faster-RCNN-original-paper}, (middle) sparse learnable proposals in~\SparseRCNN~\cite{Sparse-RCNN}, and (right)~\ProposalType~in~\ModelName.
}
\label{fig:Proposals_Comparison}
\end{figure*}

Building on the same motivation, a transformer-based approach, \colA{DEtection TRansformer (DETR)}, is introduced in~\cite{DETR}. 
DETR treats object detection as a direct set prediction problem, utilizing only 100 learned object queries as input.
With a fixed small set of learned object queries, DETR simultaneously reasons the relationships between objects and the global image context to directly output the final predictions in parallel.
However, the underlying concept of DETR necessitates that each object query interacts with the global image context.
As a result, this dense interaction not only slows down training convergence, but also prevents the establishment of a fully sparse pipeline for object detection.

To date,~\SparseRCNN~\cite{Sparse-RCNN} is the only approach to introduce fully sparse proposals for HBB detection. 
Unlike DETR, which relies on dense interactions between object queries and global image features,~\SparseRCNN~eliminates this necessity through a novel sparse learnable proposals representation.
\colC{Therein, hundreds of thousands of box candidates are replaced by a small set of learnable proposals, referred to as sparse learnable proposals, each comprising a proposal box and its associated feature.} 
The framework employs a one-to-one interaction between these proposal features and pooled \colA{Region of Interest (RoI)} features. 
This interaction is considered fully sparse, since it does not require full image context, thereby ensuring a truly sparse representation.
The adoption of sparse learnable proposals by~\SparseRCNN~offers distinct advantages over traditional dense proposals. 
Notably, it removes the need for a \colA{region proposal network (RPN)} during training and eliminates the dependency on non-maximum suppression (NMS) during inference, as overlapping dense predictions are inherently avoided.
This streamlined approach allows~\SparseRCNN~to operate with a simplified architecture, consisting solely of a backbone feature extractor, an interaction module, and regression and classification layers.
Another noteworthy work on sparse proposal approaches is presented in~\cite{Sparse_Anchoring} where a reduced set of region proposals is generated by the \colA{Sparse Anchoring Network (SAN)}. 
However, since SAN functions as an RPN within R-CNN-based detectors, it falls short of achieving the streamlined design of~\SparseRCNN.

The fully sparse interaction within~\SparseRCNN~aligns with the characteristics of SAR ship detection where ship objects are typically sparsely distributed across large areas.
This makes~\SparseRCNN~a strong candidate for adoption in SAR ship detection.
This application, however, poses unique challenges, particularly in inshore regions where background interference from waves, reflections, and nearby structures creates significant ambiguities.
Such interference often leads to an increase of false positives, reducing the reliability of detection models in complex maritime scenarios.
% Motivated by the strengths of sparse learnable proposals and the challenges specific to inshore ship detection, our research focuses on designing an improved R-CNN framework that leverages sparse learnable proposals for enhanced performance in inshore environments. 
\colC{Motivated by the strengths of sparse proposals and the challenges of inshore ship detection, we design an enhanced R-CNN framework tailored for improved performance in such environments.}

\colB{
We introduce R-Sparse R-CNN, the first extension of sparse learnable proposals for detecting arbitrarily oriented objects, tailored for SAR ship detection. The prefix “R” signifies the adaptation of Sparse R-CNN to the rotated object detection task through oriented bounding box regression.}
By embedding learnable orientation into the proposal,~\ModelName~addresses the limitations posed by the existing sparse proposal-based method in capturing object orientation. 
The ability to capture orientation plays a pivotal role in ship detection, enabling downstream tasks such as trajectory prediction and collision avoidance.
\ModelName~unifies the robustness of~\SparseRCNN~with a cost-free proposal design, while completely eliminating the need for RPN and NMS post-processing. 
This novel design enables simpler model structure and simplifies end-to-end design and training in a single shot.
The proposed model can be viewed as an extension of~\SparseRCNN~for HBB detection in~\cite{Sparse-RCNN}, inheriting its unique properties while advancing its capability to estimate object's orientation.

To improve ship detection in inshore regions, we propose~\ProposalType~(\ProposalTypeAbbr), a novel proposal concept which incorporates contextual information around the object for enriched object representation.
A comparison between the traditional proposals used in Faster R-CNN, the learnable proposals in~\SparseRCNN, and \colA{the proposed~\ProposalTypeAbbr~is shown in~\Figure~\ref{fig:Proposals_Comparison}}. 
Traditional proposals (\Figure~\ref{fig:Proposals_Comparison}-left) rely solely on learnable bounding boxes, while sparse learnable proposals (\Figure~\ref{fig:Proposals_Comparison}-middle) extend this by including learnable object features alongside the boxes. 
Advancing further, \colA{the proposed~\ProposalTypeAbbr~}(\Figure~\ref{fig:Proposals_Comparison}-right) \colC{integrate} both learnable object features and corresponding background features, offering a richer and more comprehensive object representation.
% These features interact with their corresponding counterparts extracted via the RoI pooling operation, and the resulting features are fused to form the final representation for box regression and class prediction.
\colC{These features interact with their RoI-pooled counterparts and are fused to form the final representation for box regression.}
% These one-on-one interactions and subsequent fusion of object and background features allow the model to learn the relationship between objects and their surroundings and to better distinguish ships from cluttered backgrounds, ultimately improving the detection accuracy.
\colC{These one-on-one interactions and feature fusion enable the model to capture object–background relationships, enhancing its ability to distinguish ships from cluttered scenes and improving detection accuracy.}
\colA{The proposed model is the first} sparse proposal-based detector to incorporate contextual features for improved performance, marking a novel contribution to this line of research.

In summary, the main contributions of this article are summarized as follows:
\begin{enumerate}
\item We propose~\ModelName, a novel oriented SAR ship detection model built on sparse learnable proposals.  
\ModelName~is the first to leverage sparse learnable proposals for object detection in the SAR image domain and is specifically designed to improve detection performance in inshore regions.
\colA{The proposed} model achieves a streamlined architecture by eliminating dense anchor design and avoiding both RPN training and NMS post-processing.
\item We introduce~\ProposalType~(\ProposalTypeAbbr), which leverage both ship and background features for enriched object representation. 
By capturing the relationships between these features, \colA{the proposed approach enhances} the model's contextual understanding.
Experimental results confirm that incorporating~\ProposalTypeAbbr~into the proposed model leads to a notable improvement in overall detection accuracy.

\item 
\colA{
We introduce~\DCP, a novel pooling strategy to extract both object and background features in a single operation. 
By avoiding separate pooling,~\DCP~accelerates the feature extraction process and improves detection performance. 
This improvement is attributed to its design, which extracts both features from the same Feature Pyramid Network (FPN) level, ensuring consistent feature alignment.
}
\item
\colA{
We design an~\InteractionModule~to enable one-to-one interactions between proposals and RoI features. 
This module comprises two~\InteractionHeads~for object-object and background-background interactions, followed by a~\FusionHead~that merges the resulting features.
}
\end{enumerate}

The remainder of this article is organized as follows, Section \ref{Section:Related-Work} explains in more depth the concepts of SAR ship detection and sparse learnable proposals. 
Section \ref{Section:Methodology} provides the implementation details of \colA{the proposed}~\ModelName~model. 
The experimental details are provided in Section \ref{Section:Experiment_Setup}, while the experiment results, performance comparison to other models, and the subsequent discussion are covered in Section \ref{Section:Results_Discussion}.
Finally, Section~\ref{Section:Conclusion} concludes this paper.

\section{Related Work}
\label{Section:Related-Work}

\subsection{Deep Learning for Oriented SAR Ship Detection}
\label{SubSection:CNN-SAR-Ship-Detection}

Recent progress in deep learning, particularly Convolutional Neural Networks (CNNs), has promoted their extensive use in oriented ship detection within SAR imagery~\cite{Guo_Ship_OBB_2023, YOLOv5_BiFPN, YOLO-Lite_Lighweight, YOLOv3-Tiny_RealTime, Ma_GF3, CoAM_RFIM, DCMSNN_2018, ARPN_2020, Faster_R_CNN_Ship_2021, Cascade_RCNN_ship_2024, Mask_RCNN_MultiTemporal_102019, SiS_SARShip_MaskRCNN_012023, FBR-Net_072020, CenterNet++_Ship_042021, Ship_FCOS_HRS_072021, FCOS_Ship_Large-Scale_022022}. 
Existing oriented ship detection methods typically extend detectors designed for natural images with horizontal bounding boxes (HBB) by incorporating additional parameters to represent orientation.
Based on the number of stages required and whether they use predefined anchor boxes, these detectors can be categorized into one-stage, two-stage, and anchor-free detectors.

\begin{figure*}
    \centering

\begin{tabular}{@{}b{.27\linewidth}@{\hspace{1.5em}}b{.36\linewidth}@{\hspace{1.5em}}b{.31\linewidth}@{}}
    \includegraphics[width=\linewidth]{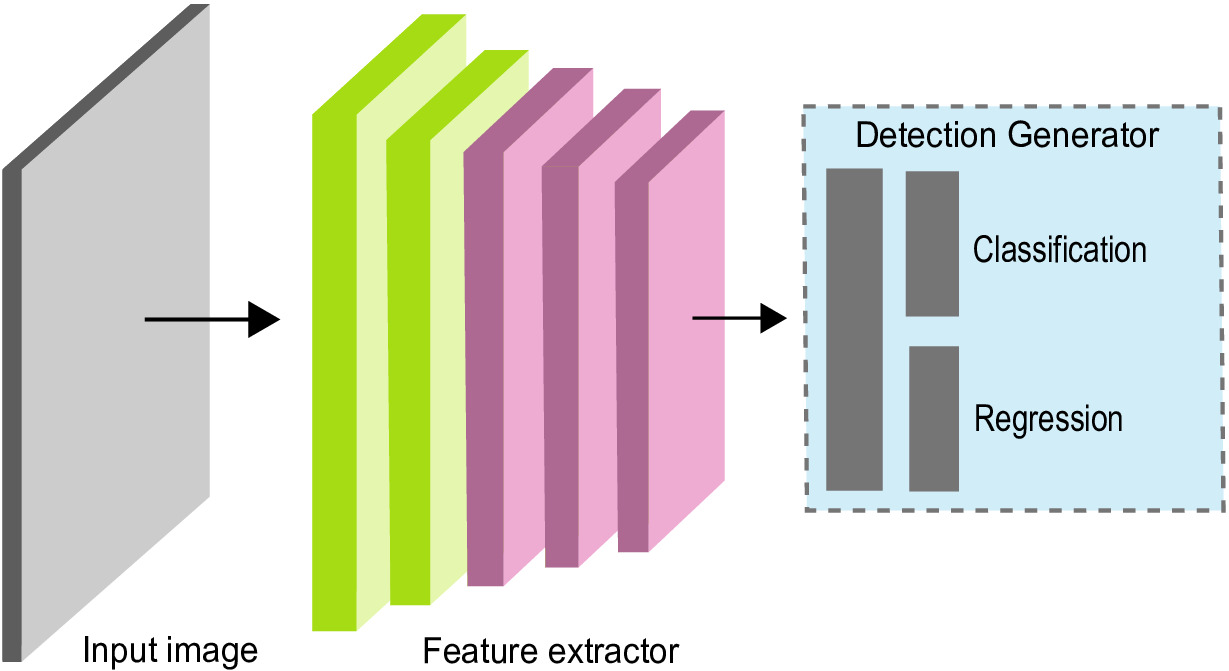} & 
    \includegraphics[width=\linewidth]{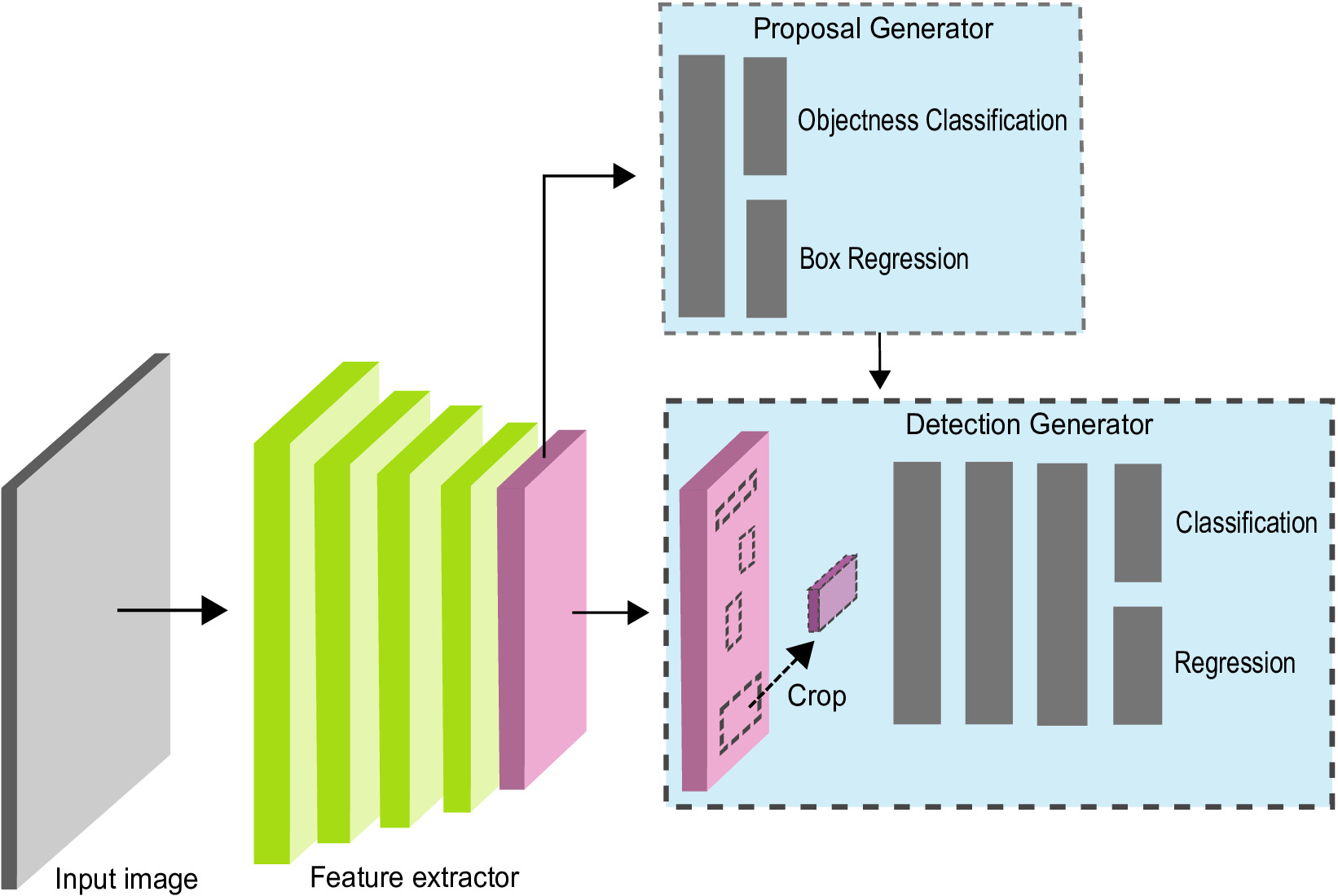} & 
    \includegraphics[width=\linewidth]{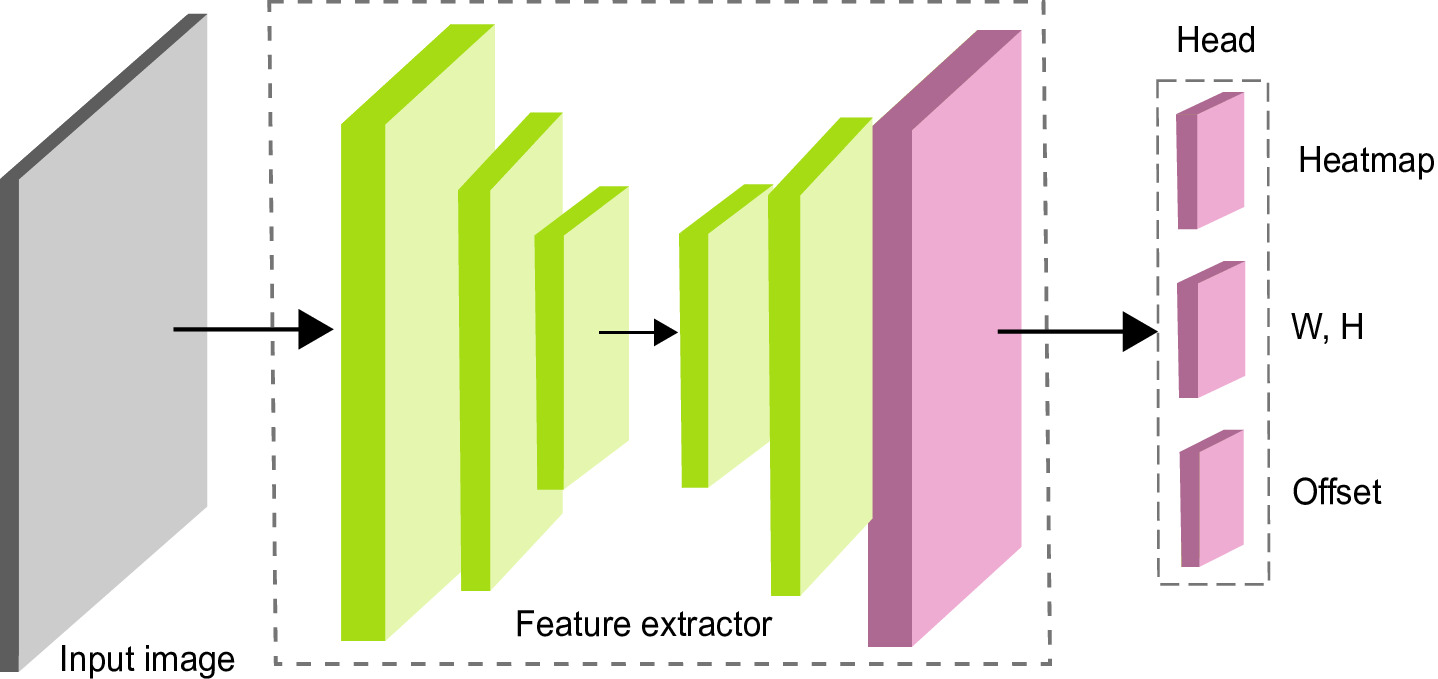}\\[\abovecaptionskip]
    \centering \small (a) One-stage detector & \centering \small (b) Two-stage detector & \centering \small \colB{(c) Anchor-free detector}
\end{tabular}
    \caption{\colB{Architecture comparison between one-stage, two-stage, and anchor-free object detector ((a) and (b) are re-drawn from~\cite{Faster-RCNN_vs_Anchor-Free}).}}
    \label{fig:One_Stage_vs_Two_Stage}
\end{figure*}

\colA{
\subsubsection{One-stage detectors}
\label{One_Stage_Detectors}
One-stage detectors (Fig.\ref{fig:One_Stage_vs_Two_Stage}a) streamline object localization and classification by performing both tasks in a single, unified step. 
These models treat detection as a dense regression and classification task by applying detection heads directly to feature maps extracted from the input image. 
The image is divided into a grid, and each grid cell predicts object class probabilities and bounding box offsets relative to predefined anchors or reference points. 
By removing the need for an explicit region proposal stage, one-stage detectors achieve faster inference speeds and lower computational complexity.
}

\colA{
You Only Look Once (YOLO)\cite{YOLO-v1} and Single Shot Detector (SSD)~\cite{SSD_paper} are two of the most widely adopted frameworks, forming the foundation for many subsequent one-stage detection methods.
Building on these frameworks, various adaptations and improvements have been proposed for oriented SAR ship detection.
For example, Yu and Shin \cite{YOLOv5_BiFPN} enhanced the YOLOv5 model by combining coordinate attention blocks with a bidirectional feature pyramid network (BiFPN) for better feature fusion. 
Ren~\EtAl~\cite{YOLO-Lite_Lighweight} used YOLO-Lite with a feature-enhanced backbone and a channel and position enhancement attention (CPEA) module to capture detailed positional information. 
Li~\EtAl~\cite{YOLOv3-Tiny_RealTime} optimized YOLOv3-Tiny for real-time SAR ship detection by tuning the prediction layer and incorporating a Convolutional Block Attention Module (CBAM).
Ma~\EtAl~\cite{Ma_GF3} adopted SSD and introduced MR-SSD, utilizing a three-resolution input from the GaoFen-3 (GF3) dataset.
}
\colC{Additionally}, Yang~\EtAl\cite{CoAM_RFIM} further improved SSD's robustness to scale variations by incorporating a coordinate attention and receptive field enhancement module.

\colA{
\subsubsection{Two-stage detectors}
\label{Two_Stage_Detectors}
A two-stage detector (Fig. \ref{fig:One_Stage_vs_Two_Stage}b) first generates region proposals and then refines them for final object detection.
In the first stage, anchor boxes of various scales and aspect ratios are placed over feature maps. 
A Region Proposal Network (RPN) refines these anchors via bounding box regression to generate object proposals.
In the second stage, RoI features corresponding to these proposals are extracted and passed through classification and regression heads to predict object categories and refine box coordinates.
}

\colA{
The R-CNN family of detectors—including Faster R-CNN~\cite{Fast-RCNN-original-paper}, Region-based Fully Convolutional Network (R-FCN)~\cite{R-FCN}, Cascade R-CNN~\cite{Cascade_RCNN}, and Mask R-CNN~\cite{Mask_RCNN}—serves as the foundation for most two-stage object detectors.
In the context of oriented SAR ship detection, several methods have adapted and extended these architectures.
Jiao~\EtAl~\cite{DCMSNN_2018} introduced a model named DCMSNN, which is built upon the Faster R-CNN framework to address the challenges of multiscale and multisource SAR ship detection. 
Rather than relying on a single feature map, their approach connects feature maps from top to bottom and generates proposals from each fused map.
Zhao~\EtAl~\cite{ARPN_2020} introduced the Attention Receptive Pyramid Network (ARPN) to enhance the detection of multiscale ships in SAR images. 
This model improves performance by strengthening the relationships among non-local features and refining the information across different feature maps.
Ke~\EtAl~\cite{Faster_R_CNN_Ship_2021} introduced an enhanced Faster R-CNN framework, incorporating deformable convolutional kernels to better capture the geometric transformations of shape-varying ships.
Chai~\EtAl~\cite{Cascade_RCNN_ship_2024} presented an improved Cascade R-CNN model designed to detect small-sized ship targets within complex backgrounds.
Additionally, the works presented in~\cite{Mask_RCNN_MultiTemporal_102019} and~\cite{SiS_SARShip_MaskRCNN_012023} demonstrate the utilization of Mask-RCNN for SAR ship detection.
}

\colA{
\subsubsection{Anchor-free detectors}
\label{Anchor_Free_Detectors}
Anchor-free methods offer an alternative to traditional anchor-based object detectors by eliminating the need for predefined anchor templates. 
Instead, these models detect objects by directly regressing keypoints, regions, or pixels that indicate object presence. 
Notable examples include CenterNet~\cite{CenterNet}, CornerNet~\cite{CornerNet}, and FCOS~\cite{FCOS}. 
}
\colB{
CenterNet (Fig. \ref{fig:One_Stage_vs_Two_Stage}c) locates object centers through heatmap regression and predicts bounding box dimensions via offset estimation. 
}
\colA{
CornerNet detects objects by identifying top-left and bottom-right corners, which are then grouped to form bounding boxes. 
FCOS formulates detection as a per-pixel regression task, where each location predicts distances to the object’s boundaries and its class label.
These anchor-free paradigms have been adapted for oriented SAR ship detection. 
Fu~\EtAl~\cite{FBR-Net_072020} introduced the Feature Balancing and Refinement Network (FBR-Net) with an Attention-Guided Balanced Pyramid (ABP) for multiscale feature balancing, alongside a Feature-Refinement (FR) module for accurate localization.
Cui~\EtAl~\cite{Spatial_Shuffle_Group_CenterNet_012021} proposed a CenterNet-based model incorporating a Spatial Shuffle-Group Enhancement (SSE) module to strengthen semantic feature extraction and reduce false positives.
Guo~\EtAl~\cite{CenterNet++_Ship_042021} extended CenterNet with a feature refinement module and feature pyramid fusion to improve small ship detection. 
Finally, Sun~\EtAl~\cite{Ship_FCOS_HRS_072021} enhanced FCOS with a Category-Position (CP) module for improved position regression in complex scenes, while Zhu~\EtAl~\cite{FCOS_Ship_Large-Scale_022022} further optimized FCOS with focal loss, regression refinement, and Complete Intersection over Union (CIoU) loss for large-scale SAR ship detection.
}

\subsection{Sparse Learnable Proposals}
\label{SubSection:Overview-Sparse-Learnable-Proposals}

The key idea of sparse learnable proposals, first introduced in~\SparseRCNN~\cite{Sparse-RCNN}, is the substitution of hundreds of thousands of box candidates (anchors) with a small set of \textit{N} learnable proposals. 
In typical R-CNN architectures like Faster R-CNN, the Region Proposal Network (RPN) generates dense anchors as candidate regions, learning objectness and location refinement for proposal boxes. 
\SparseRCNN, however, replaces dense anchors with learnable proposals that are randomly initialized and iteratively refined via their interactions with pooled RoIs from feature maps.
This design eliminates the dependency on dense anchors design and RPN training as well as obviates the need for NMS post-processing commonly due to densely overlapping predictions.
Thus, the sparse proposals concept enables~\SparseRCNN~to achieve a streamlined design.

\begin{figure}[htb]
\includegraphics[width=0.48\textwidth,center, trim={0 0 0 0}]{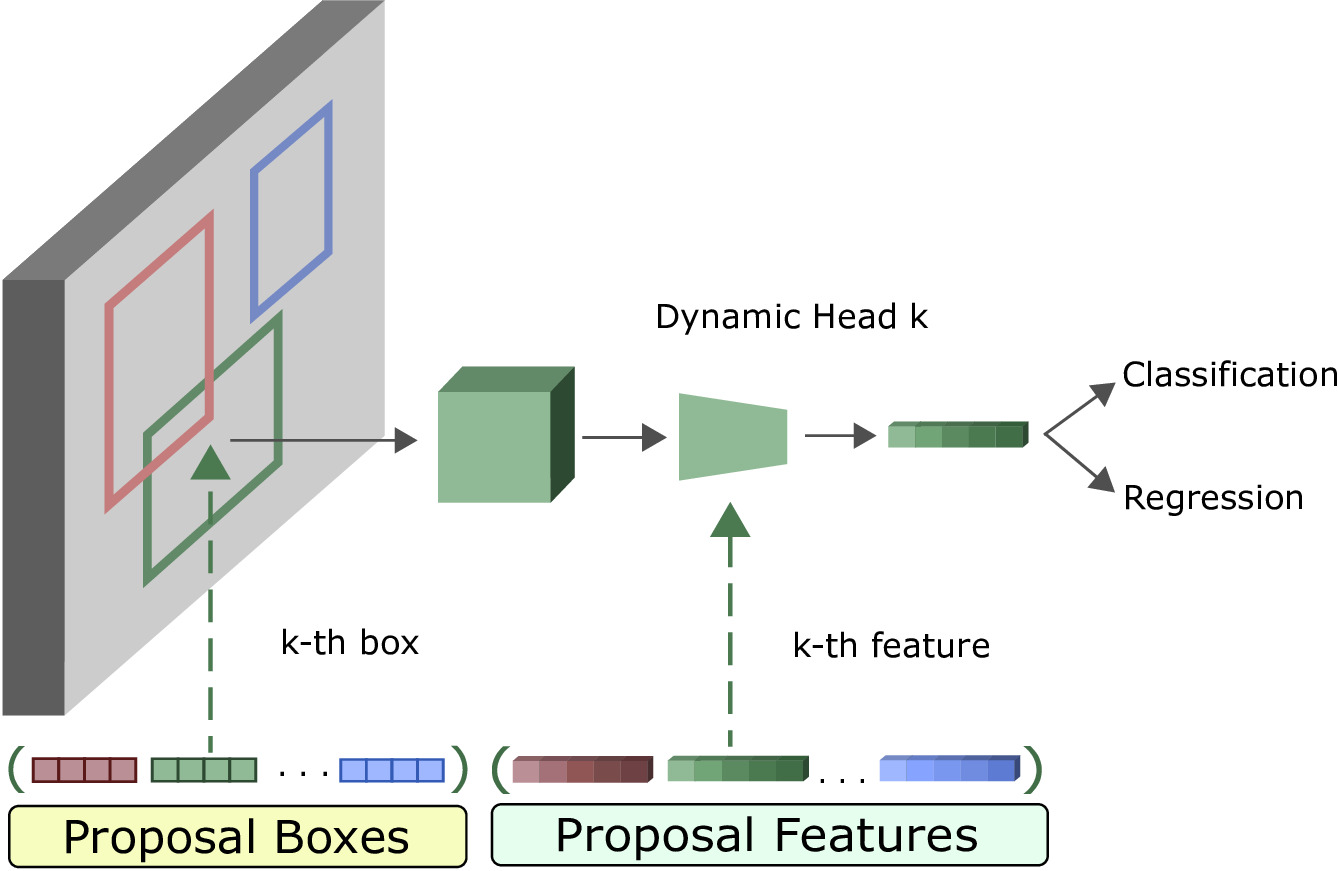}
\caption{
    Illustration of sparse learnable proposals within~\SparseRCNN. Each proposal \colC{consists of} proposal box and learnable proposal feature. The proposal feature will interact with pooled RoI feature via its dynamic head to refine the box and predict the corresponding class (re-drawn from~\cite{Sparse-RCNN}).
}
\label{fig:SparseRCNN_Flowchart}
\end{figure}

From a technical standpoint, learnable proposals represent an improved version of traditional proposal boxes enhanced with additional learnable features referred to as proposal features. 
% The learned proposal boxes, in essence, are the statistical distribution of potential object locations within the training set and can be interpreted as an initial approximation of regions likely to contain objects, independent of the input.
\colC{
The learned proposal boxes represent the statistical distribution of object locations in the training set and serve as input-independent initial estimates of regions likely to contain objects.
}
On the other hand, proposals generated by RPN are directly influenced by the specific input image and provide a rough estimation of object locations.
While the proposal box provides a brief and explicit representation of objects, it offers only coarse localization and misses essential details such as object pose and shape.
Proposal features, however, encode rich instance characteristics, allowing each proposal to embed not only box-related parameters but also unique instance features that capture these missing details. 
The illustration of~\SparseRCNN~pipeline which leverages the sparse learnable proposals is given in Fig.~\ref{fig:SparseRCNN_Flowchart}.

In its original implementation~\cite{Sparse-RCNN}, the interaction between proposal and RoI features is realized through a~\DynInstInteraction~head.
For \(N\) proposal boxes, the model begins by applying the RoIAlign~\cite{Mask_RCNN} operation to extract features for each proposal box, referred to as RoI features.
Each RoI feature is processed through its dedicated head for object localization and classification. 
Each head is specifically conditioned on the corresponding proposal feature.
Each RoI feature \( f_i(S \times S, C)\) interacts with the corresponding proposal feature \( p_i(C) \) to refine its representation, resulting in the final object feature \( O_i(C) \). 
Specifically, this feature interaction mechanism is realized through two consecutive \( 1 \times 1 \) convolutions with ReLU activation,  where the weights for these convolutions are dynamically derived from the corresponding proposal features $p_i$.
Due to the nature of this interaction, the mechanism is referred to as~\DynInstInteraction.
The pseudo-code of~\DynInstInteraction~is provided in Algorithm~\ref{alg:Dynamic_Instance_Interaction}.
The term \(bmm\) in Algorithm~\ref{alg:Dynamic_Instance_Interaction} refers to batched matrix multiplication.

\begin{algorithm}
\caption{Dynamic Instance Interaction}
\label{alg:Dynamic_Instance_Interaction}
\begin{algorithmic}[1]
    \State \textbf{Input:} Proposal feat. \( p(N, C) \), RoI feat. \( f(N, S \times S, C)\)
    \State Calculate dynamic parameters $P \sim Linear_{1}(p)$
    \State Extract first parameters $P_1 \sim P[:, :length/2]$
    \State Extract second parameters $P_2 \sim P[:,length/2:]$
    \State Calculate $F_1 \sim ReLU(Norm(bmm(f, P_1)))$
    \State Calculate $F_2 \sim ReLU(Norm(bmm(F_1, P_2)))$
    \State Calculate output features $O \sim Linear_{2}(F_2)$
    \State \textbf{Output:} \(O\)
\end{algorithmic}
\end{algorithm}

\section{Methodology}
\label{Section:Methodology}

This section provides an in-depth explanation of \ModelName, introducing its core concept alongside \colA{the proposed~\ProposalType}. 
Additionally, we present the overall architecture of
~\ModelName, which consists of a backbone network and a dedicated~\DetectionHead, as illustrated in~\Figure~\ref{fig:Background-Aware-Sparse-Proposals}.
Finally, we discuss the training objectives, inference process, and the loss functions utilized by the model.

\subsection{Rotated Learnable Proposals}
\label{SubSection:Rotated-Learnable-Proposals}

\begin{figure}[htb!]
\includegraphics[width=0.2\textwidth,center]{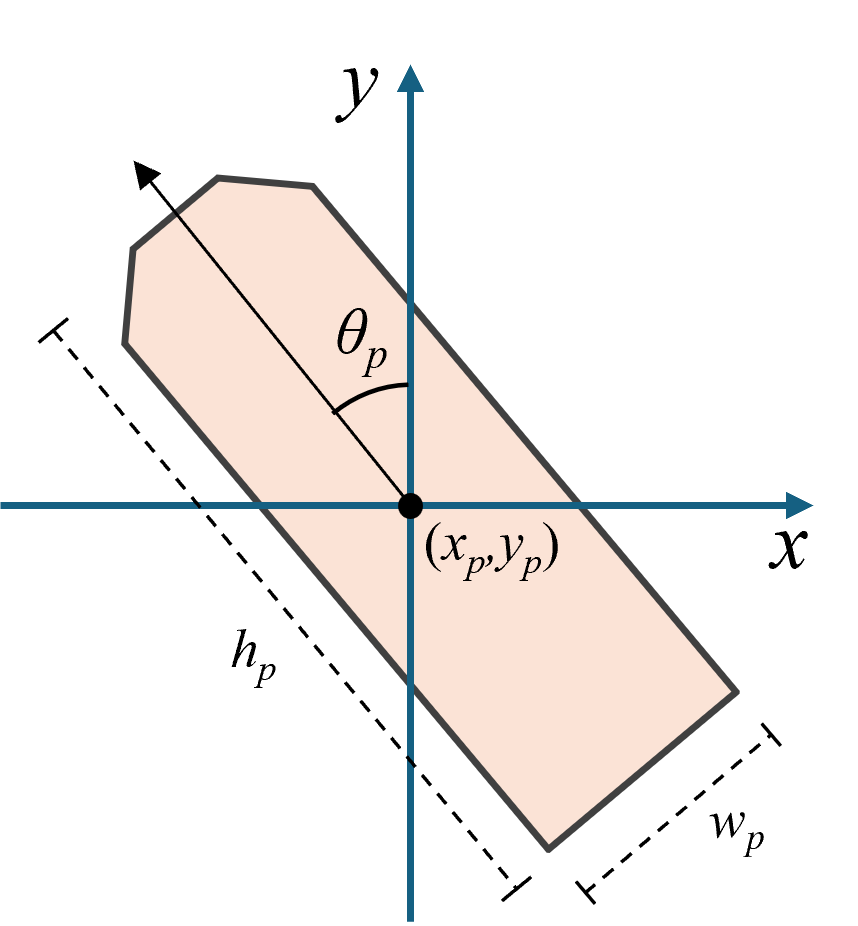}
\vspace{0.05cm}
\caption{Oriented bounding box representation used in proposed~\ModelName.}
\label{fig:Ship-Coordinate}
\end{figure}

We extend the original concept of sparse learnable proposals, initially designed for detecting axis-aligned objects, to facilitate the detection of oriented objects.
To achieve this, we introduce additional parameters to represent orientation, defining each proposal box by a 5-dimensional vector of $(x_p$, $y_p$, $w_p$, $h_p$, $\theta_p)$. 
An illustration of these parameters, applied to represent ship coordinates, is shown in Fig.~\ref{fig:Ship-Coordinate}.
The ($x_p$, $y_p$) represents the center point of the proposal box while ($w_p$, $h_p$, $\theta_p$) respectively represent the width, height, and orientation. 

To accommodate the embedding of the orientation parameter $\theta_p$, we redesigned the baseline model~\cite{Sparse-RCNN} with two significant adjustments.
First, in the RoI pooling stage, instead of using standard RoIAlign~\cite{Mask_RCNN} that originated for HBB, we used Rotated RoIAlign (R-RoIAlign) on the feature maps output by the backbone. 
This is intended to better capture the feature of the targets enclosed by rotated proposal boxes. 
For simplicity, we refer to this operation as ``RoI pooling'' throughout the remainder of this paper.
Secondly, we modified the regression layer to output five parameters instead of four. These parameters represent the offsets for rotated box attributes: location \((\delta_{x}, \delta_{y})\), width \((\delta_{w})\), height \((\delta_{h})\), and orientation \((\delta_{\theta})\). 

\subsection{Background-Aware Proposals}
\label{SubSection:Background-Aware-Proposals}

The interaction between proposal features and RoI features within~\SparseRCNN~framework allows the model to establish a one-to-one reasoning mechanism between each proposal and its corresponding RoI feature. 
This pairwise interaction mechanism provides a dedicated attention space for each object proposal, enabling the network to iteratively refine its understanding on each candidate for improved detection performance.
Building upon this object-object interaction mechanism, we hypothesize that incorporating background-background interactions and fusing the resulting interactions from both object-object and background-background reasoning could further enhance detection performance. 
By allowing the model to jointly reason about objects and their corresponding background contexts through feature fusion, it can learn the typical background patterns associated with specific objects. 
This enriched contextual understanding would enable the model to better discriminate objects from their surroundings, improving detection accuracy and robustness in complex scenarios.

\begin{figure*}[t]
\includegraphics[width=1.0\textwidth,center, trim={0 0.9cm 0 0}]{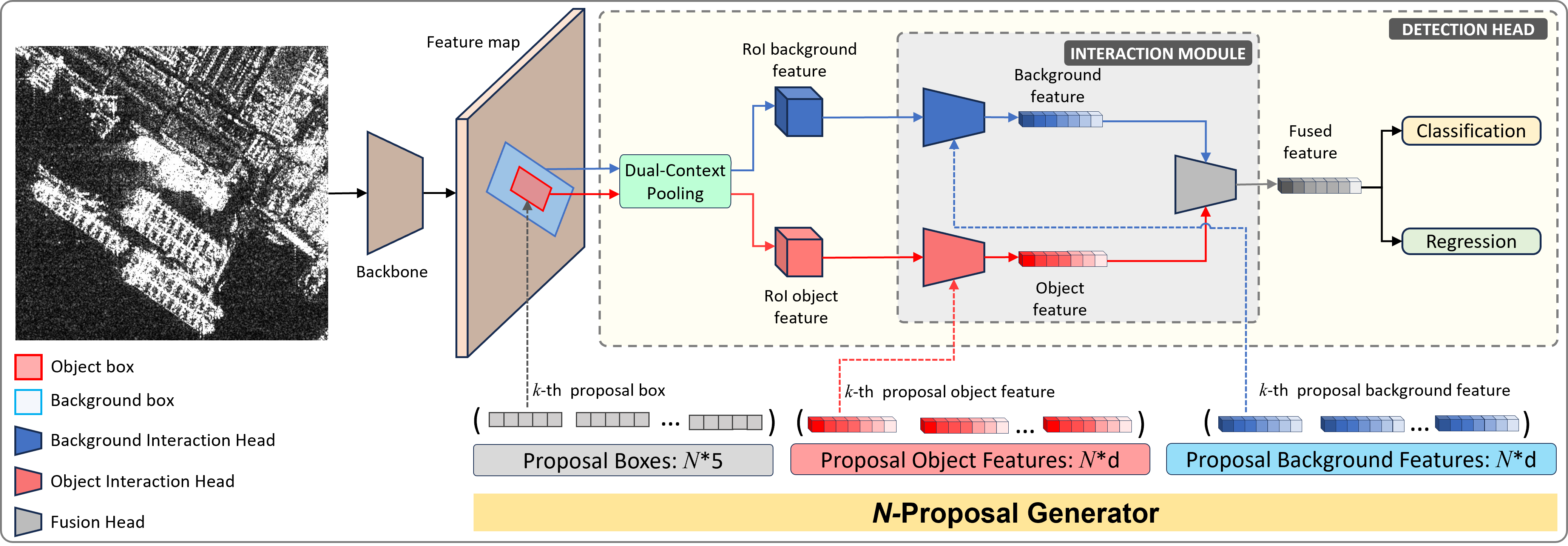}
\vspace{0.05cm}
\caption{
    Architecture of the proposed~\ModelName~pipeline featuring background-aware rotated sparse proposals.
    proposal-RoI interactions are realized via~\InteractionHeads~followed by a~\FusionHead.
}
\label{fig:Background-Aware-Sparse-Proposals}
\end{figure*}

\colD{
To facilitate background-background interactions, we propose~\ProposalType, an extension of traditional learnable proposals that incorporates contextual features, referred to as proposal background features.
Hence,~\ProposalType~are composed of three elements: proposal boxes, object features, and background features.
The proposal background features are specifically designed to interact with the background features extracted from regions surrounding the pooled RoI features, termed RoI background features.
The interaction mechanism between proposal features and RoI features, along with the overall structure of the proposed model, is illustrated in~\Figure~\ref{fig:Background-Aware-Sparse-Proposals} and further detailed in~\Section~\ref{Interaction-Module}.
}

Introducing background-background interactions to the pipeline allows the model to learn what “typical” background looks like within each RoI context. 
This contextual sensitivity is particularly valuable for SAR ship detection tasks in maritime environments, where background elements (e.g., sea surface textures or port infrastructure) vary significantly but may closely resemble object shapes. 
By leveraging~\ProposalType, \colA{the proposed model integrates} contextual information into object features to better distinguish ships from their surroundings, thus reducing false positives and enhancing detection reliability in complex scenes.

\subsection{Backbone}
\label{Backbone}
\begin{figure}[H]
\includegraphics[width=0.50\textwidth,center, trim={-0.3cm 0 -0.3cm 0}]{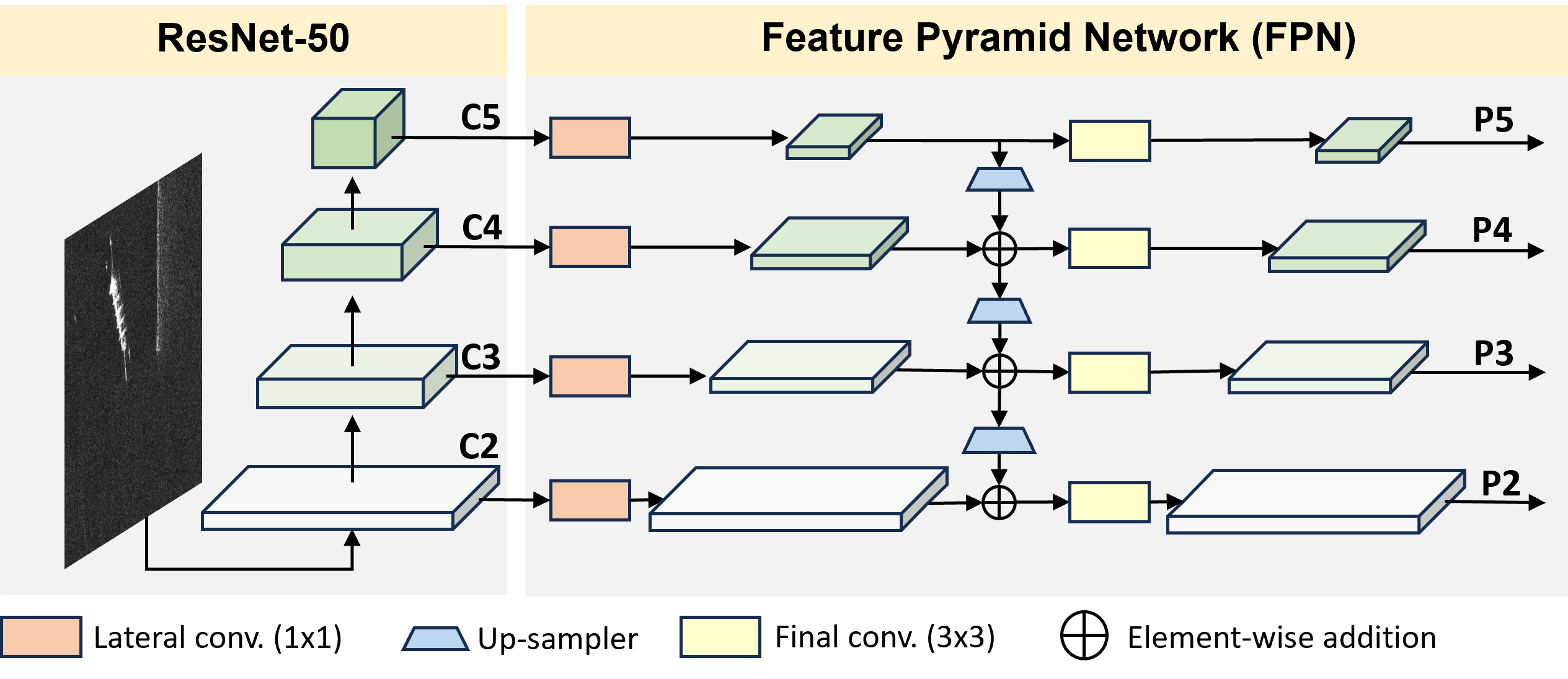}
\caption{
    Architecture of ResNet-50-FPN backbone.
}
\label{fig:ResNet50_FPN}
\end{figure}

~\ModelName~employs a ResNet-50 backbone with a Feature Pyramid Network (FPN) for multi-scale feature fusion, referred to as ResNet-50-FPN as depicted in Fig.~\ref{fig:ResNet50_FPN}.
Initially, residual blocks of ResNet-50 (\textit{res2}, \textit{res3}, \textit{res4}, \textit{res5}) output feature maps C2, C3, C4, and C5, which are then passed to the FPN. 
Afterwards, to standardize their varying channel dimensions (2048, 1024, 512 and 256 respectively), a 1×1 lateral convolution is applied to align all these maps to 256 channels. 
The aligned maps are subsequently fused in a top-down manner, with smaller maps upsampled to match the larger ones. Finally, a final 3×3 convolution refines the fused maps, producing outputs P2, P3, P4, and P5.

\subsection{\DetectionHead}
As illustrated in~\Figure~\ref{fig:Background-Aware-Sparse-Proposals}, the ~\DetectionHead~consists of three main parts, i.e.,~\DCP,~\InteractionModule, and classification and regression layers.

\subsubsection{Dual-Context Pooling}
\label{SubSection:Dual-Context-Pooling}
The~\ModelName~exploits two key interactions: between RoI and proposal background features, and between RoI and proposal object features.
Capturing these interactions requires a pooling scheme that jointly extracts both RoI background and object features.
A straightforward and commonly adopted solution involves applying separate pooling operations for the object and background regions. 
However, this approach introduces two significant drawbacks.
First, the dual pooling operation inherently doubles the computational cost, leading to a substantial increase in processing time.
Second, due to the larger size of background boxes compared to object boxes, background and object features may be extracted from different levels of the feature pyramid.
Specifically, smaller regions use deeper, high-resolution feature maps, while larger regions rely on coarser, low-resolution maps. 
Consequently, this discrepancy causes resolution and detail misalignment between features, leading to suboptimal interactions between object and background features.
Such interactions hinder the performance of subsequent processes that rely on the accurate fusion of these features.

\begin{figure*}[ht]
\includegraphics[width=0.74\textwidth,center]{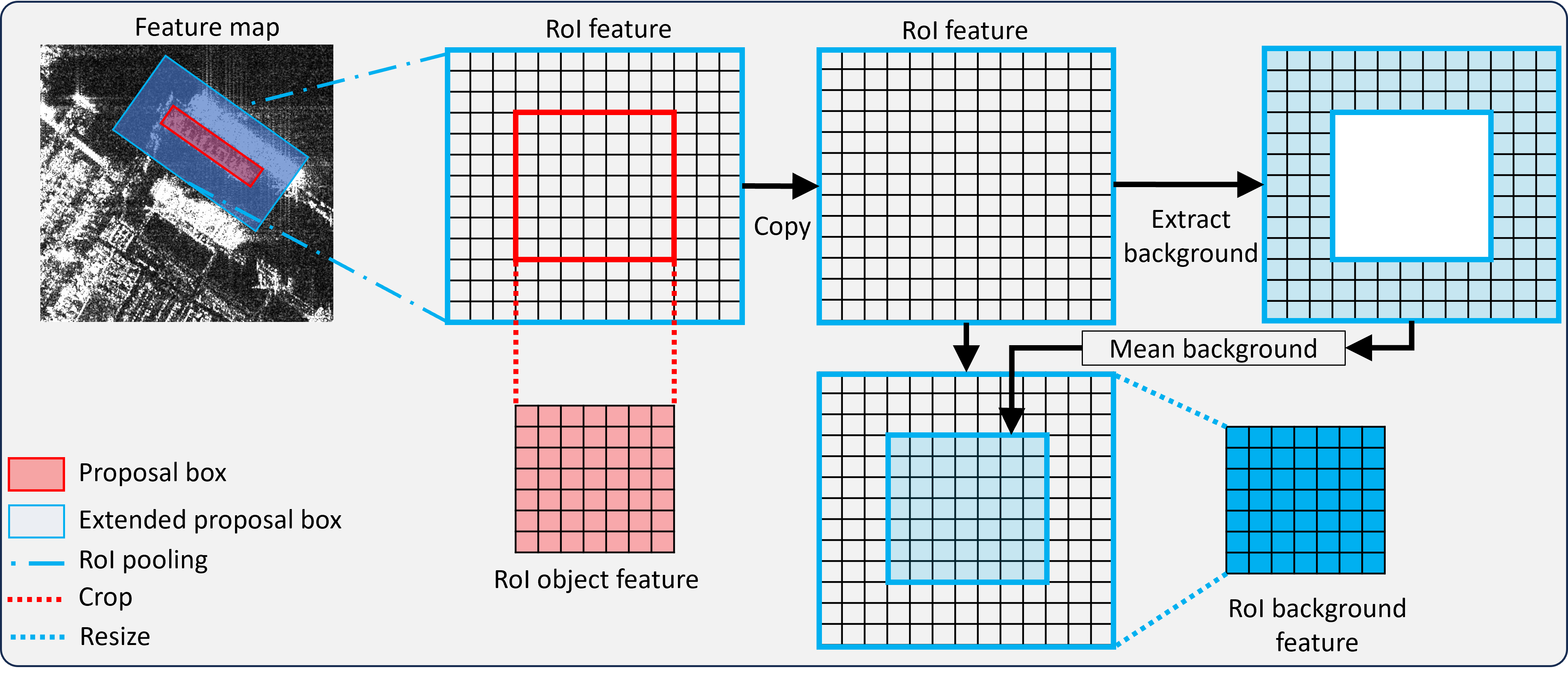}
\caption{Extraction of object and background features via the proposed~\DCP~mechanism.}
\label{fig:DCP}
\end{figure*}

To overcome the aforementioned issues, we propose the~\DCP~(DCP) mechanism which captures both background and object features in a single, unified operation. 
By extracting features from the same level of the feature pyramid, DCP ensures consistent feature alignment and strengthens the relationship between extracted object and background features. 
The detailed process of DCP is illustrated in~\Figure~\ref{fig:DCP}.

The DCP module is designed to extract both object and background features from a given object proposal box. 
This is achieved through a two-step process: (1) applying a pooling operation over an extended RoI on feature map $F$  to capture features containing both object and background features $F_{roi}$, and (2) center-cropping $F_{roi}$ to extract object feature $F_{roi}^{obj}$, while processing the remaining feature to obtain the background feature \colD{$F_{roi}^{bg}$}.
In our implementation, both object and background features are set to have dimensions of \(7 \times 7\) pixels, following commonly adopted settings in~\cite{Faster-RCNN-original-paper} and~\cite{Sparse-RCNN}.

Specifically, given a proposal box $P_{box}\sim(x_p, y_p, w_p, h_p, \theta_p)$, RoI pooling operation is performed on assigned feature map $F$ over an extended region approximately 1.86 times the size of $P_{box}$ (designated ${P_{box}\uparrow}$), resulting in a $13 \times 13$ RoI feature $F_{roi}$. 
From $F_{roi}$, $7 \times 7$ pixels center-cropping is applied to extract RoI object feature $F_{roi}^{obj}$.
\colA{The extension factor $\alpha = 1.86$ corresponds to the ratio \(13/7\)}, which is chosen to enable direct center-cropping on $13 \times 13$ \colD{$F_{roi}$} to extract \( 7 \times 7 \) $F_{roi}^{obj}$.
This specific factor avoids additional interpolation.
Such interpolation would otherwise be required when center-cropping a \(7 \times 7\) region from a \(14 \times 14\) pooled feature map, as in the case of using an \(\alpha\) of 2.0. 
In that case, the operation involves fractional pixel locations, necessitating interpolation to estimate values at sub-pixel positions.
\colA{
This choice of $\alpha$ is further supported by experiments in Section~\ref{sub:Exp_DCP}, where \( \alpha = 13/7 \) achieves better detection performance than other practical ratios \(9/7\), \(11/7\), \(15/7\), and \(17/7\).
}

To extract the RoI background feature $F_{roi}^{bg}$, we first replace the central $7 \times 7$ values of $F_{roi}$ with the mean of its surrounding background $\mu^{\text{bg}}$. 
\colB{
This operation removes object-specific features and replaces them with a statistical representation of the background, ensuring the resulting features accurately represent the background.
}
Finally, the modified $F_{roi}$ is downsized to $7 \times 7$, resulting in \colD{$F_{roi}^{bg}$}.
For clarity, the operations involved in DCP mechanism are provided in~\Equation~(\ref{eqn:F_RoI}) to ~\Equation~(\ref{eqn:F_bg_RoI})

\begin{equation}
\label{eqn:F_RoI}
F_{roi} = \text{RoIPool}(F, P_{box}\uparrow)_{13\times13},
\end{equation}
\begin{equation}
\label{eqn:F_obj_RoI}
F^{obj}_{roi} = \text{CenterCrop}(F_{roi})_{7\times7},
\end{equation}
\begin{equation}
\label{eqn:F_bg_RoI}
F^{bg}_{roi} = \text{Resize}(\text{ReplaceCenter}(F_{roi},\mu^{\text{bg}})_{7 \times 7})_{7 \times 7}.
\end{equation}

% The proposed~\DCP~approach effectively captures both background and object features in a single pass, eliminating the need for separate pooling processes.
% Our experimental results in~\Section~\ref{sub:Exp_DCP} demonstrate that this approach achieves superior detection accuracy compared to separately pooling the background and object features, highlighting the effectiveness of the approach.

\colD{
The extracted $F^{obj}_{roi}$ and $F^{bg}_{roi}$ features are then fed into an interaction module, enabling interaction with the proposal object feature $F^{obj}_{pro}$ and background object feature $F^{bg}_{pro}$, respectively. 
}
Detailed architecture and functionality of this module are provided in the next section.

\subsubsection{\InteractionModule}
\label{Interaction-Module}
The~\InteractionModule~consists of two specialized~\InteractionHeads~and one~\FusionHead.

\paragraph{\InteractionHeads}
\label{Interaction-Heads}
The first~\InteractionHead~facilitates interaction between RoI object features and proposal object features, while the second handles interaction between RoI background features and proposal background features. 
For clarity, these heads are referred to as~\ObjectHead~and~\BackgroundHead, respectively.
In our implementation, these two heads share the same architecture, namely the interaction head architecture of~\cite{Sparse-RCNN}.

\begin{figure*}[t]
\includegraphics[width=0.9\textwidth,center, trim={0 0 0 0}]{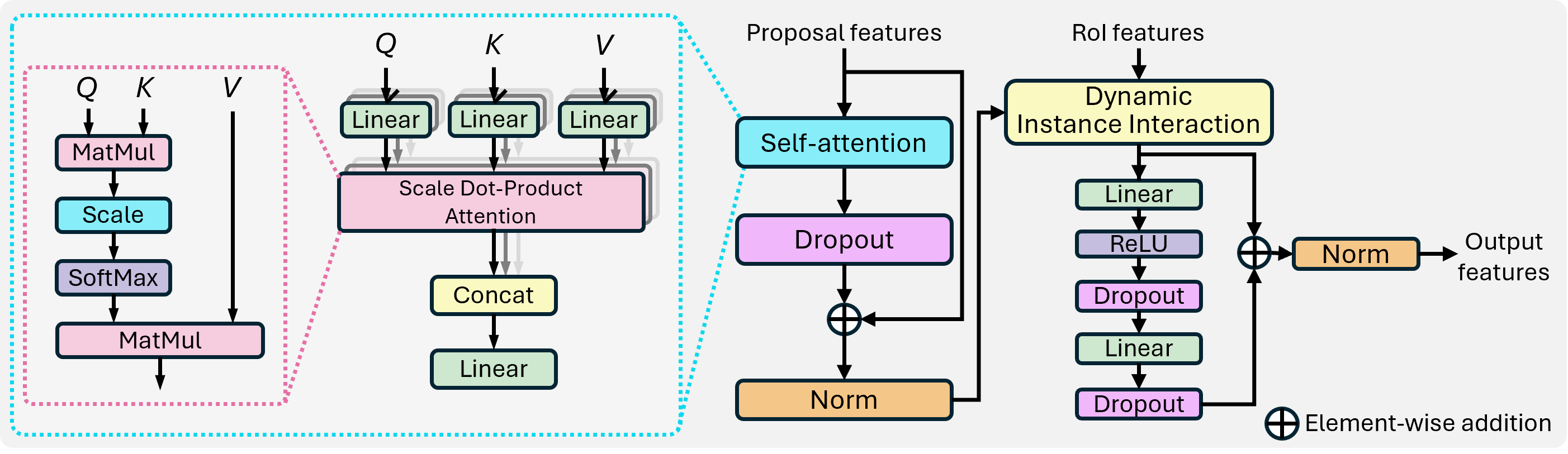}
\vspace{0.05cm}
\caption{Architecture of the~\InteractionHead~featuring self-attention module and~\DynInstInteraction~mechanism.}
\label{fig:Interaction-Head}
\end{figure*}

Feature interaction within each head employs a~\DynInstInteraction~(DII) as detailed~in Algorithm~\ref{alg:Dynamic_Instance_Interaction}. 
Further details on the structure of~\InteractionHead~are illustrated in Fig.~\ref{fig:Interaction-Head}.
\colC{Following the approaches in~\cite{Sparse-RCNN} and~\cite{Relation_Networks}, we apply a self-attention operation using a multi-head attention block on the proposals before passing them to the DII layer. This operation enables the model to reason about relationships among objects and among backgrounds within the set of proposals.}
The general formulation of attention mechanism is given as 

\begin{equation}
\label{eqn:Attention}
\text{Attention}(Q, K, V) = \text{softmax} \left( \frac{Q K^T}{\sqrt{d_k}} \right) V.
\end{equation}
Where $Q = W_Q X$, $K = W_K X$, and $V = W_V X$ are the query, key, and value matrices respectively. $W_Q$, $W_K$, and $W_V$ are learnable weight matrices, where $d_k$ is the dimension of the feature vector.
In the case of self-attention, the same input \( X \) is used for generating \( Q \), \( K \), and \( V \), allowing each element in \( X \) to attend to all other elements in the input sequence, thereby capturing contextual relationships within \( X \) itself.

For the~\BackgroundHead, which interacts RoI background features \( F^{bg}_{roi} \) and proposal background features \( F^{bg}_{pro} \), a dropout operation is applied to the output of the self-attention on \( F^{bg}_{pro} \). 
The resulting features are then residual-added to the original \( F^{bg}_{pro} \) and passed through a normalization layer.
The normalized features then interact with RoI features via the DII. 
\begin{equation}
\label{eqn:F_bg_pro}
F^{bg}_{pro} = \text{Norm}(F^{bg}_{pro} + \text{Dropout}(\text{Attention}(F^{bg}_{pro})))
\end{equation}
\begin{equation}
\label{eqn:Dynamic-Interaction}
F^{bg} = DII(F^{bg}_{roi}, F^{bg}_{pro}) 
\end{equation}
Finally, the output from~\Equation~\ref{eqn:Dynamic-Interaction}~is passed through two consecutive linear layers for further refinement and alignment with the initial input dimension.

As previously mentioned, the~\ObjectHead~shares the same architecture as the~\BackgroundHead; thus,~\Equation~(\ref{eqn:F_bg_pro}) and~\Equation~(\ref{eqn:Dynamic-Interaction}) apply to this head as well. 
The only difference is that the ~\colD{\ObjectHead~takes} proposal object features $F^{obj}_{pro}$ and RoI object features $F^{obj}_{roi}$ as inputs, instead of $F^{bg}_{pro}$ and $F^{bg}_{roi}$, and outputs the object features $F^{obj}$.

\paragraph{Fusion Head}
\label{Fusion-Head}
The proposed~\FusionHead~combines object features $F^{obj}$ and background features $F^{bg}$ produced by the~\InteractionHeads~to generate a unified representation that captures the relationship between objects and their context. 
To achieve this, we leverage a cross-attention mechanism as in Eq.~(\ref{eqn:Attention}), where $F^{obj}$ act as the queries and keys, and $F^{bg}$ serve as the values. 
This setup allows each object feature to selectively attend to relevant background information, creating a context-aware representation that strengthens object-background relationship.

The architecture of \colA{the proposed}~\FusionHead~is simple, incorporating only a dropout layer and normalization following the cross-attention operation to stabilize and refine the fused features.
\begin{equation}
\label{eqn:FusionHead}
F_{fuse} = \text{Norm}(\text{Dropout}(\text{Attention}(F^{obj},F^{bg}))).
\end{equation}
In addition to cross-attention, we experimented with alternative fusion strategies, such as direct feature addition and concatenation.
However, experimental results showed that cross-attention approach yields better accuracy, making it the preferred choice for capturing detailed contextual dependencies between objects and their background.

\subsubsection{Classification and Regression Layers}
\label{Subsubsection:Classfication-Regression-Module}

At the final stage, the~\DetectionHead~takes the fused features $F_{fuse}$ as input to classify proposals and regress their bounding boxes.
To this purpose, we use a straightforward structure consisting of a linear layer followed by normalization and ReLU activation for both the regression and classification layers. 
Specifically, the classification layer outputs class logits, while the regression layer outputs offsets for the proposal box parameters \(\delta_{x}\), \(\delta_{y}\), \(\delta_{w}\), \(\delta_{h}\), and \(\delta_{\theta}\).
The initial proposal box parameters are then updated by following equation:
\begin{equation}
\label{eqn:box-update}
\begin{gathered}
    \hat{x} = x_p + \delta_x \cdot w_p \cdot \cos\theta_p + \delta_y \cdot h_p \cdot \sin\theta_p,\\
    \hat{y} = y_p + \delta_x \cdot w_p \cdot \sin\theta_p + \delta_y \cdot h_p \cdot \cos\theta_p,\\
    \hat{w} = w_p \cdot e^{\delta_w},\\
    \hat{h} = h_p \cdot e^{\delta_h},\\
    \hat{\theta} = \theta_p + \delta_\theta.
\end{gathered}
\end{equation} 

According to~\cite{Sparse-RCNN, DETR, Cascade_RCNN}, a stack of~\DetectionHeads~was shown to enhance model performance. 
Following this strategy, \colA{the proposed}~\ModelName~employs the same approach by stacking multiple~\DetectionHeads~to iteratively refine the proposals.
Specifically, the updated box parameters computed in~\Equation~\ref{eqn:box-update}, along with the $F^{obj}$ and $F^{bg}$ features output by~\InteractionHeads~of the current head serve as inputs to the subsequent head.

\subsection{Training and Inference}
\label{SubSection:Training-Inference}

\subsubsection{Training Details}
\label{SubSubSection:Training-Details}
The training stage of~\ModelName~closely resembles the second stage of Faster R-CNN~\cite{Faster-RCNN-original-paper}.
In this stage, the feature maps produced by the backbone, specifically P2 through P5, are used to extract RoI features through the DCP operation.
\colC{
The key difference is that, unlike Faster R-CNN, which directly feeds RoI features to the classification and regression layers,~\ModelName~takes a different approach. 
It first interacts these features with proposal features through the~\InteractionModule~to produce fused features.
}
The fused \colC{features} are then passed to the classification layer to compute class logits and to the regression layer to determine offset values, which map the proposal boxes to the predicted boxes. 
These predictions are subsequently used to compute the loss, enabling  the updating of model parameters via backpropagation.

\subsubsection{Inference Details}
\label{SubSubSection:Inference-Details}
The inference process in~\ModelName~is simple and efficient. Given an input image, the model directly outputs \textit{N} bounding boxes with corresponding confidence scores. These predictions are used as-is for evaluation, without requiring any NMS post-processing.

\subsection{Loss Function}
In ~\ModelName~training, two types of loss are computed: matching loss and training loss. 
The matching loss measures the differences between generated proposals and the ground truth. 
Specifically, proposals are matched to ground truth objects in a one-to-one manner based on the matching cost, where a lower cost indicates a closer match.
The matching cost is formulated as:
\\
\begin{equation}
\mathcal{L} = \lambda_{\text{cls}} \cdot \mathcal{L}_{\text{cls}} + \lambda_{\text{L1}} \cdot \mathcal{L}_{\text{L1}} + \lambda_{\text{iou}} \cdot \mathcal{L}_{\text{iou}}.
\end{equation}
\\
In this formulation, $\mathcal{L}_{\text{cls}}$ denotes the focal loss, which measures the discrepancy between the predicted classifications and the actual category labels. 
The terms $\mathcal{L}_{\text{L1}}$ and $\mathcal{L}_{\text{iou}}$ represent the L1 loss and \colA{the Intersection over Union (IoU) loss computed between rotated bounding boxes}, respectively. 
The coefficients $\lambda_{\text{cls}}, \lambda_{\text{L1}}, \lambda_{\text{iou}}$ are used to weight each component of the loss. 
The calculation of training loss uses the same formulation as matching loss but only for matched proposal–ground truth pairs.
Finally, the total loss is calculated as the sum of all pairs, normalized over the number of matched pairs in the training batch.

\section{Experimental Setup}
\label{Section:Experiment_Setup}

\subsection{Dataset}
\ModelName~is trained and evaluated (separately) on two publicly available datasets, SSDD\cite{SSDD_Dataset} and RSDD-SAR \cite{RSDD-SAR}. 
\subsubsection{SSDD}
SSDD was the first publicly available dataset for oriented ship detection in SAR images and has been released in two versions. The 2021 version is used for this work, which includes uniform OBB annotations. This dataset comprises of 1,160 images, containing a total of 2,456 ships. It spans a wide range of resolutions, image sizes, and polarization modes. 
SSDD is widely used as a standard benchmark for evaluating the performance of ship detection methods in SAR imagery.
\subsubsection{RSDD-SAR}
The RSDD-SAR dataset is specifically designed for OBB annotations in SAR images. It \colC{consists of} 7,000 images and includes a total of 10,263 ships. This dataset features a range of resolutions, polarization modes, and imaging techniques. It serves as a valuable resource for testing the generalization capability of the proposed model and for evaluating ship detection performance based on OBB annotations.

Both the SSDD and RSDD-SAR datasets include inshore and offshore test sets, which are essential for evaluating the model’s performance across different environments and backgrounds. The specific parameters of SSDD and RSDD-SAR datasets are detailed in Table \ref{table:RSDD_vs_SSDD_Properties}.

\begin{table}[htbp]

\caption{SSDD and RSDD-SAR dataset parameters}
\renewcommand{\arraystretch}{1.5} % Adjust the value to increase the spacing
\centering
\begin{tabular}{p{1.5cm}p{3.0cm}p{3.5cm}}
	\hline
	Dataset & SSDD & RSDD-SAR \\ \hline
	Sensors        & RadarSat-2, TerraSAR-X, Sentinel-1    & TerraSAR-X, Gaofen-3      \\ 
	Polarization   & HH, HV, VH, VV                        & HH, HV, VH, VV, DH, DV    \\ 
	Resolution     & 1--15                                 & 2--20                     \\ 
	Scenes         & Inshore, offshore                     & Inshore, offshore         \\ 
	Image size     & 160--668                              & 512 $\times$ 512          \\ 
	Image count    & 1,160                                 & 7,000                     \\ 
	Ship count     & 2,456                                 & 10,263                    \\ 
	Annotation     & HBB, OBB, polygon                     & OBB                       \\ \hline
\end{tabular}
\label{table:RSDD_vs_SSDD_Properties}
\end{table}

\subsection{Evaluation}
% read practical : https://learnopencv.com/mean-average-precision-map-object-detection-model-evaluation-metric/

To quantitatively evaluate the performance of~\ModelName, the \colA{average precision (AP)} metric from the Microsoft Common Objects in Context (MS COCO)~\cite{MS_COCO} evaluation framework is utilized. 
The AP is formulated as:

\begin{equation}
AP = \int_{0}^{1} \text{P}(\text{R}) \, d(\text{R}).
\end{equation}
The Precision (P) and Recall (R) are defined as follow:

\begin{equation}
\text{Precision} = \frac{N_{TP}}{N_{GT}}
\end{equation}

\begin{equation}
\text{Recall} =  \frac{N_{TP}}{N_{DET}},
\end{equation}
where \( N_{TP} \) stands for true positives, indicating the number of correctly identified positive samples. \( N_{GT} \) refers to the number of ground truth boxes while \( N_{DET}\) denotes the number of predictions made by the model. 

The AP metric evaluates model performance at different IoU thresholds to represent true positive predictions.
In this work, we utilize AP, AP$_{50}$, and AP$_{75}$ as the evaluation metrics. 
AP is computed as the mean precision across 10 IoU thresholds ranging from 0.5 to 0.95 in steps of 0.05, providing a comprehensive measure of detection accuracy. 
AP$_{50}$ represents precision at an IoU threshold of 0.5, serving as a lenient criterion, while AP$_{75}$ employs a stricter IoU threshold of 0.75, emphasizing higher localization accuracy.

\subsection{Hyperparameters and Environment}
\label{Hyperparameters_and_Environment}
The backbone of~\ModelName~was set up with weights pre-trained on ImageNet \cite{ImageNet}, \colD{whereas the newly introduced layers, including the proposal features,  were initialized using Xavier initialization \cite{Xavier_Initialization}}. 
The model was then trained using the Adam optimizer with an initial learning rate of \(7.5 \times 10^{-5}\) for 150 epochs, with momentum set to 0.9, weight decay set to 0.0001, and L2 norm gradient clipping. 
A warm-up strategy was implemented for the first 50 iterations, using one-third of the initial learning rate. 
The learning rate was further reduced by a factor of 10 at epochs 130 and 140. 
\colA{
The loss weight coefficients $\lambda_{\text{cls}}$, $\lambda_{\text{L1}}$, and $\lambda_{\text{iou}}$ were set to 2.0, 5.0, and 2.0, respectively, following the baseline,~\SparseRCNN, in~\cite{Sparse-RCNN} to ensure fair comparison. 
}
\colB{Data augmentation included random horizontal flipping with a probability of 0.5 and random resizing of the shortest edge within a range of 128 to 800 pixels.}
Training was performed with a batch size of 8 on two NVIDIA RTX 2080 GPUs (16GB combined memory), while comparisons with other models were conducted on a single NVIDIA RTX 3090 GPU.
\colC{
The software setup used Detectron2 version 0.3, built on PyTorch 2.4, running on a Linux system (Rocky Linux 8.9), and hosted by the Advanced Computing Research Centre HPC at the University of Bristol.
}

%##################################################################
% =================== RESULTS AND DISCUSSION ======================
%##################################################################

\section{Results and Discussion}
\label{Section:Results_Discussion}
In this section, we provide experimental results on~\ModelName.
We first conduct an ablation study using the SSDD-SAR dataset, as its smaller image count allows for faster experimentation. 
The optimal hyperparameters obtained are then applied to train a separate model on the RSDD-SAR dataset.
We then evaluate models' performance against state-of-the-art models on both SSDD and RSDD-SAR datasets.
\colC{Although the model trained on RSDD-SAR uses hyperparameters tuned solely on SSDD, consistent performance is observed across both datasets.
This consistency, later detailed in~\Table~{\ref{table:performance_comparison_SSDD}}~and~\Table~{\ref{table:performance_comparison_RSDD}}~and visually supported by~\Figure~\ref{fig:Visual_Comparison_SSDD}~and~\Figure~\ref{fig:Visual_Comparison_RSDD}, demonstrates the effectiveness and generalizability of the hyperparameters.}

\subsection{Proposal Configurations}
% In this section, we conducted experiments to evaluate the impact of proposal configurations on model performance. 
% Specifically, we varied the number of proposals and employed different strategies for initializing the proposal boxes. 
% This analysis provides insights into how these factors influence overall model performance.
\colC{
In this section, we conduct experiments to evaluate the impact of proposal configurations on model performance. 
Specifically, we varied the number of proposals and employed different strategies for the initialization of the proposal boxes. 
}

\subsubsection{Number of Proposals}
The number of proposals is a key determinant of model performance for proposal-based detectors. 
For instance, Faster R-CNN~\cite{Faster-RCNN-original-paper} initially employed 300 proposals which was later increased to 2000 for performance gains. 
Similarly, Cascade R-CNN~\cite{Cascade_RCNN} uses up to 1000 proposals while~\SparseRCNN~leverages 100 to 500 proposals.

\begin{table}[t]
\centering
\caption{Effect of Number of Proposals}
\label{table:Exp_Number_of_Proposal}
\renewcommand{\arraystretch}{1.2} % Adjust the value to increase the spacing
\setlength{\tabcolsep}{2.0pt}
\centering
\begin{tabular}{>{\centering\arraybackslash}m{2.5cm}>{\centering\arraybackslash}m{1.2cm}>{\centering\arraybackslash}m{1.2cm}>{\centering\arraybackslash}m{1.7cm}>{\centering\arraybackslash}m{1.5cm}}
	\hline
	Proposals& 
	AP\textsubscript{50}& 
	Model size (M)& 
	Training time (h)&
	FPS \\\hline
	
	100 & 0.927 & 185.25 & 4h & 13 \\ 
	300 & 0.929 & 185.50 & 8h & 5\\ 
	500 & 0.931 & 185.71 & 11h & 3\\ \hline
\end{tabular}
\raggedright

\end{table}

The impact of the number of proposals on model performance is presented in Table~\ref{table:Exp_Number_of_Proposal}. 
Increasing the number of proposals marginally enhances performance, but at the cost of longer training times and reduced inference speed. 
% The results also demonstrate that using more proposals does not significantly increase model size.
% This is because the increase in proposals only adds parameters for storing the 5-dimensional proposal boxes and 256-dimensional proposal features, which are minimal compared to the overall model size.
\colC{
The model size is minimally affected, as each proposal contains only a 5-dimensional box and $2 \times 256$ parameters for object and background features—negligible compared to the total model size.
}

Considering the above, the 100-proposal configuration is set as the default for the model.
However, it should be noted that the number of proposals limits the maximum detectable objects and should be adjusted based on the application. For instance, using the 100 proposals would be inappropriate for detecting hundreds of ships within a single image.

\subsubsection{Initialization of Proposal Boxes}
% In general, the performance of proposal-based detectors is heavily dependent on the initial state of box parameters. 
In general, the performance of proposal-based detectors is highly sensitive to the initial configuration of box parameters.
In this experiment, we evaluate three different initialization strategies:
\begin{itemize}
\item Center: All proposal boxes are initialized at the image center, with width and height set to $\frac{1}{4}$ and $\frac{1}{2}$ of the image size, respectively, and orientation fixed at $\frac{\pi}{4}$.
\item Random: Proposal box parameters are initialized randomly using a Gaussian distribution.
\item Grid: Proposal boxes are initialized in a regular grid across the image, following the strategy used in G-CNN~\cite{GCNN}.
\end{itemize}

\begin{table}[htbp]
\centering
\caption{Experiment on Different Box Initialization Methods}
\label{table:Exp_Pro_Box_Initialization}
\renewcommand{\arraystretch}{1.2} % Adjust the value to increase the spacing
\setlength{\tabcolsep}{2.0pt}
\begin{tabular}{>{\centering\arraybackslash}m{2.5cm}>{\centering\arraybackslash}m{1.2cm}}
	\hline
	Initialization method& 
	AP\textsubscript{50}\\\hline
	
	Center & 0.927 \\
	Random & 0.921 \\ 
	Grid & 0.919 \\ \hline
\end{tabular}

\end{table}

According to Table~\ref{table:Exp_Pro_Box_Initialization}, the model demonstrates minimal sensitivity to initialization strategies.
A slight variation of 0.8\% on $\text{AP}_{50}$ indicates that \colC{the initial position, size, and orientation of the proposal boxes have a negligible impact on overall performance}.
This robustness can be attributed to the dynamic proposal refinement mechanism of the~\DetectionHeads, which iteratively update the boxes through proposal-object interaction regardless of the boxes' initial condition. 
We chose ``Center'' as the default configuration for the rest of the experiments.

\subsection{Effect of stacking~\DetectionHeads}
Previous works in~\cite{Cascade_RCNN}~and~\cite{Sparse-RCNN}~have shown that stacked detection heads can significantly improve the detection performance.
Building on this, we adopt a similar strategy by stacking a series of~\DetectionHeads~and evaluate its effectiveness in improving detection accuracy.

\begin{table}[htbp]
\centering
\caption{Model Performance Across Different Number of \\Stacked Heads}
\label{table:Exp_Stacked_Modules}
\renewcommand{\arraystretch}{1.2} % Adjust the value to increase the spacing
\setlength{\tabcolsep}{2.0pt}
\centering
\begin{tabular}{>{\centering\arraybackslash}m{2.5cm}>{\centering\arraybackslash}m{1.2cm}>{\centering\arraybackslash}m{1.2cm}>{\centering\arraybackslash}m{1.7cm}>{\centering\arraybackslash}m{1.5cm}}
	\hline
	Heads Stacked& 
	AP\textsubscript{50}& 
	Model size (M)& 
	Training time (h)&
	FPS \\\hline
	
	2 & 0.605 & 79.75 & 2.42 & 30 \\ 
	4 & 0.906 & 132.45 & 3.32 & 18\\ 
	6 & 0.927 & 185.25 & 4.47 & 13\\
	8 & 0.901 & 238.05 & 6.59 & 10\\\hline
\end{tabular}

\raggedright
\end{table}

\Table~\ref{table:Exp_Stacked_Modules} highlights that stacking more~\DetectionHeads~consistently improves the model accuracy.
This improvement stems from the iterative refinement mechanism of the stacked design, where each successive module refines the proposals and feature representations from its predecessor, resulting in a progressively more robust model.

We observe that stacking beyond six heads results in a slight drop in accuracy (from 0.927 to 0.901), while continuing to increase model size and reduce inference speed. 
% Accordingly, we adopt a six-module configuration as the default setting to balance accuracy and efficiency.
Accordingly, we choose a configuration with six stacked~\DetectionHeads~as the default to balance accuracy and efficiency.

\subsection{Effectiveness of~\DCP}
\label{sub:Exp_DCP}
% In this section, we demonstrate the effectiveness of proposed~\DCP~(DCP) mechanism, which simultaneously extracts both RoI object features and RoI background features to facilitate RoI-proposal interaction. 
% We highlight its effectiveness compared to separately pooling object and background features, termed as ``Separate Pooling', and provide the results in~\Table~\ref{table:Exp_DCP}.
\colC{In this section, we demonstrate the effectiveness of the proposed~\DCP~(DCP) mechanism, which jointly extracts object and background features from RoIs to facilitate proposal--RoI interaction. 
We compare it against a naive approach that separately pools these features, referred to as ``Separate Pooling,'' with results reported in Table~V.}

\begin{table}[htbp]
%\centering
\caption{Experiment on Dual-Context Pooling}
\label{table:Exp_DCP}
\renewcommand{\arraystretch}{1.2} % Adjust the value to increase the spacing
\setlength{\tabcolsep}{2.0pt}
\centering
\begin{tabular}{>{\centering\arraybackslash}m{2.5cm}>{\centering\arraybackslash}m{1.2cm}>{\centering\arraybackslash}m{1.5cm}}
	\hline
	Pooling& 
	AP\textsubscript{50}& 
	FPS \\\hline
	
	DCP & 0.927 & 13 \\
        Separate Pooling & 0.906 & 12 \\ \hline
\end{tabular}

\end{table}

\colA{
\begin{figure*}
    \centering
    \begin{tabular}{@{}p{.35\linewidth}@{\hspace{0.5em}}p{.35\linewidth}@{}}
        \includegraphics[width=\linewidth]{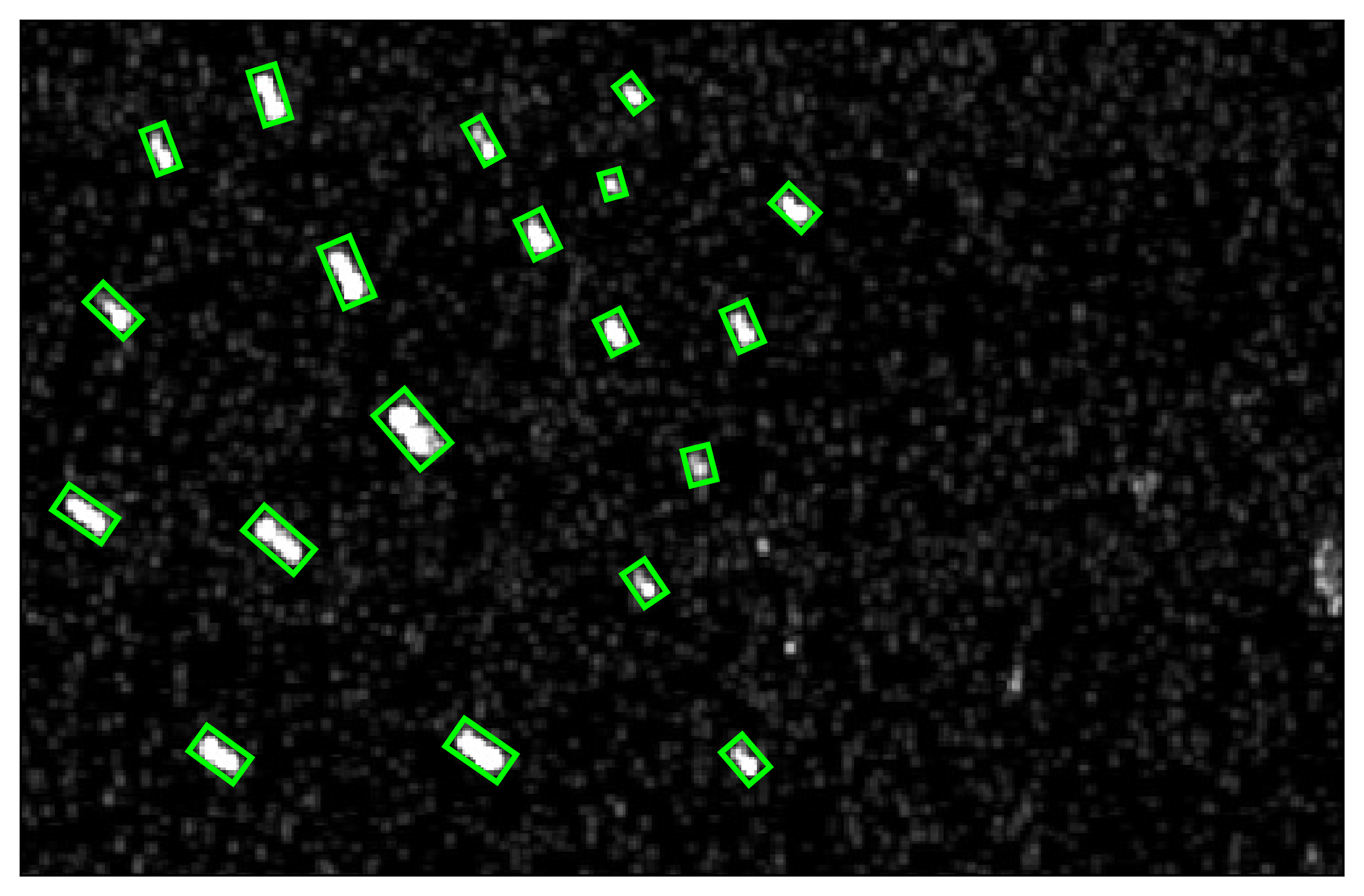} &
        \includegraphics[width=\linewidth]{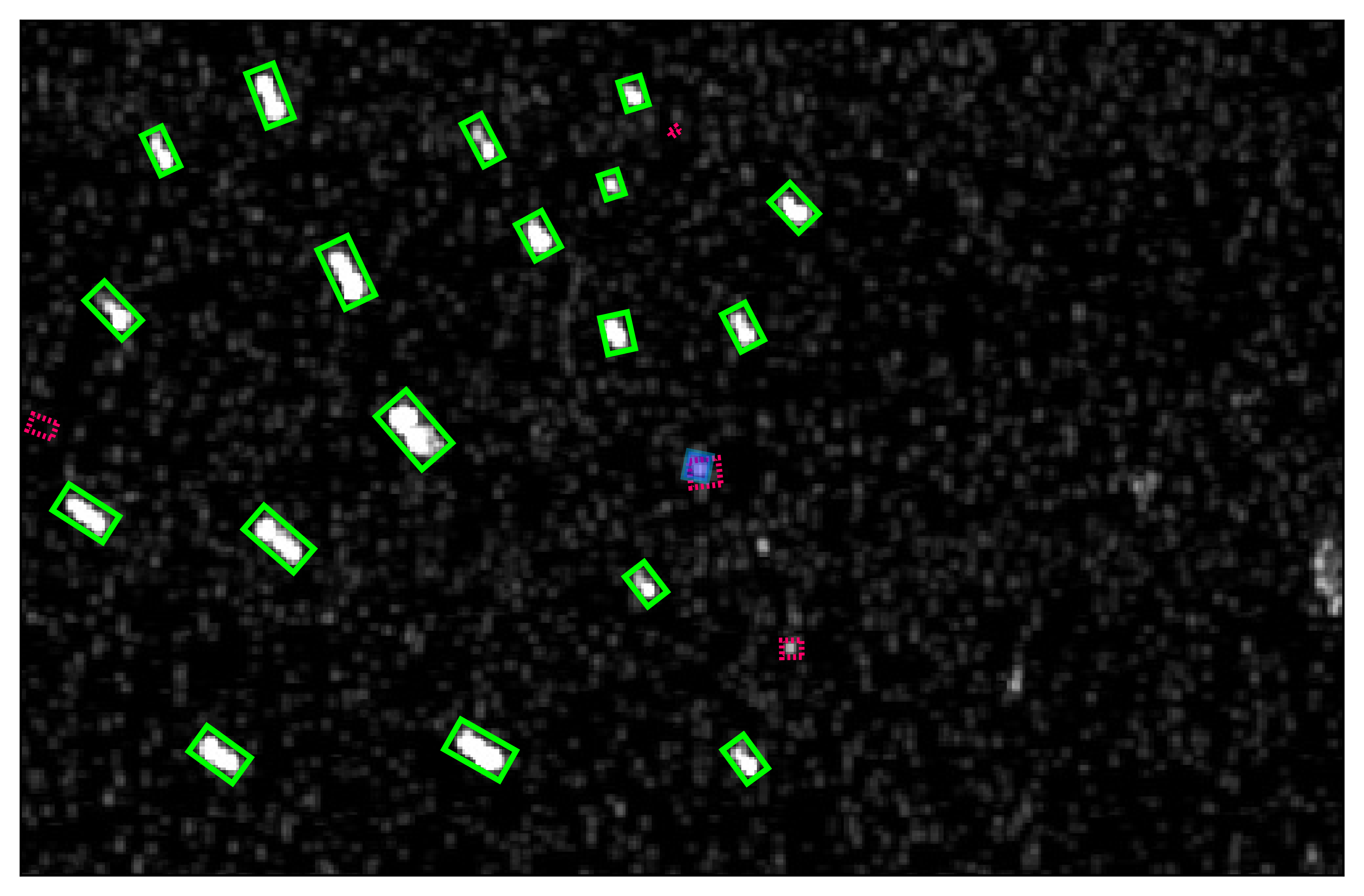} \\
        \centering\small (a) \textit{Dual-Context Pooling} & \centering\small (b) Separate Pooling
    \end{tabular}

    \vspace{1ex}  % spacing between rows

    % --- Second Row: 3 images ---
    \begin{tabular}{@{}p{.23\linewidth}@{\hspace{0.5em}}p{.23\linewidth}@{\hspace{0.5em}}p{.23\linewidth}@{}}
        \includegraphics[width=\linewidth]{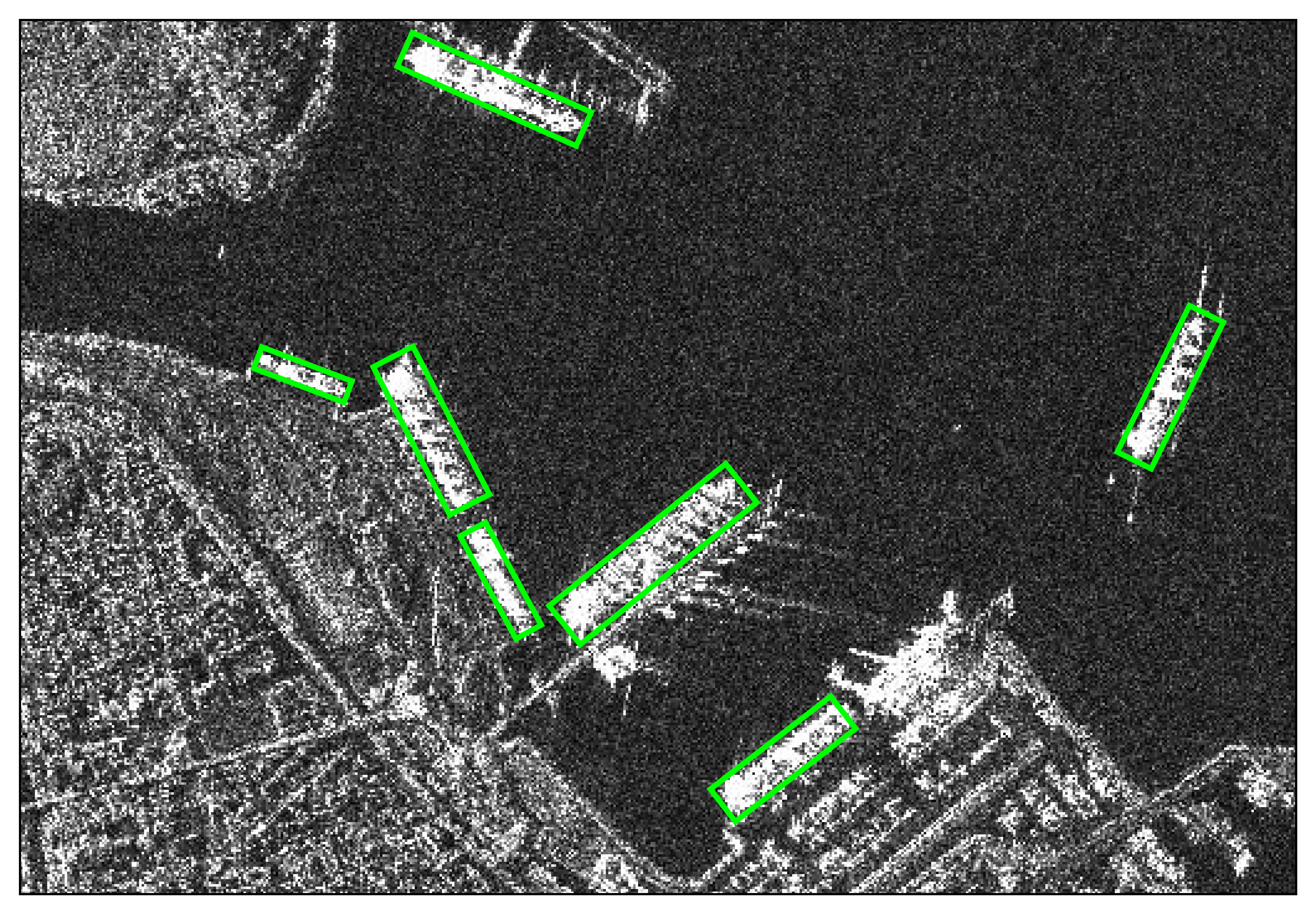} &
        \includegraphics[width=\linewidth]{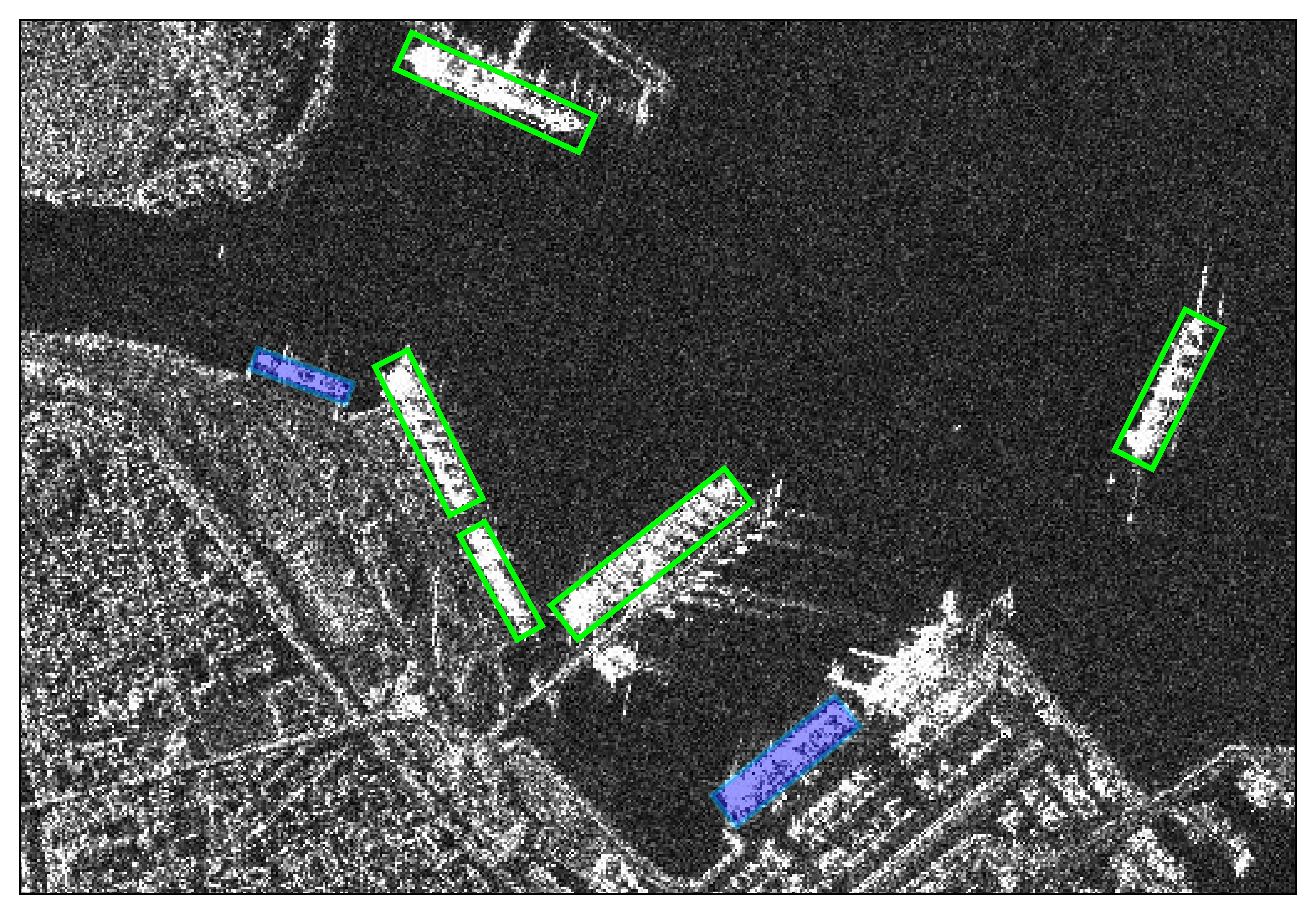} &
        \includegraphics[width=\linewidth]{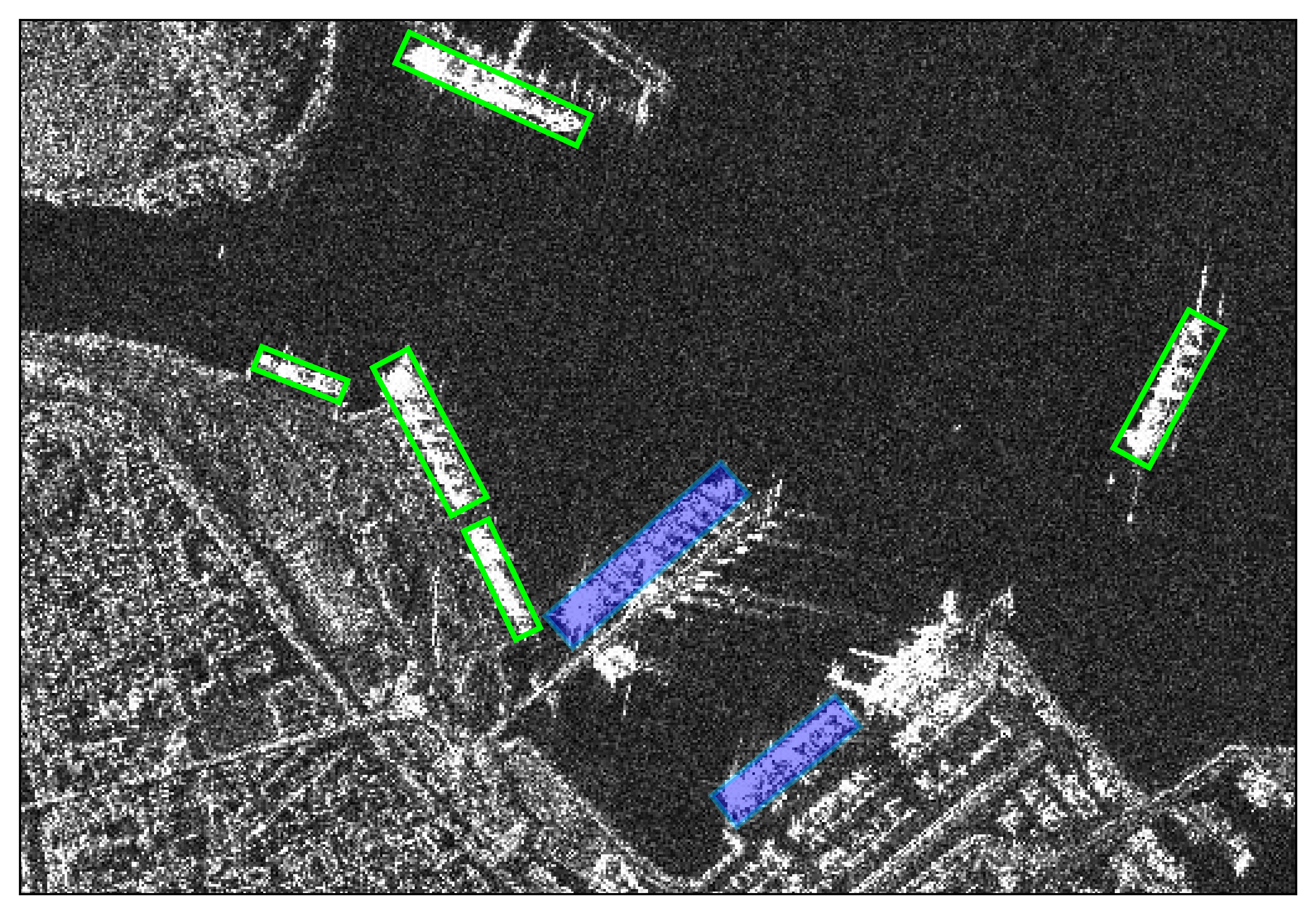} \\
        \centering\small (c) Cross-Attention & \centering\small (d) Addition & \centering\small (e) Multiplication
    \end{tabular}

    \vspace{1ex}  % spacing between rows

    % --- Third Row: 2 images ---
    \begin{tabular}{@{}p{.35\linewidth}@{\hspace{0.5em}}p{.35\linewidth}@{}}
        \includegraphics[width=\linewidth]{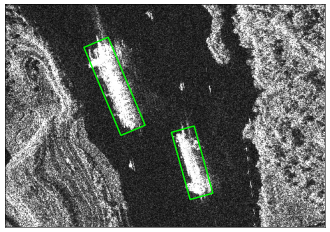} &
        \includegraphics[width=\linewidth]{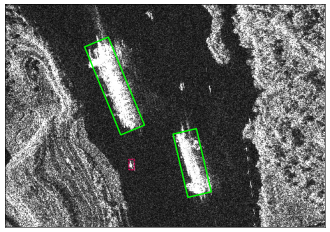} \\
        \centering\small (f) Background-Aware Proposals & \centering\small (g) Sparse Proposals
    \end{tabular}

    \vspace{0.0ex}  % spacing 
    
    \begin{tabular}{@{}p{.7\linewidth}}
        \includegraphics[width=\linewidth, trim={0 0 0 0}]{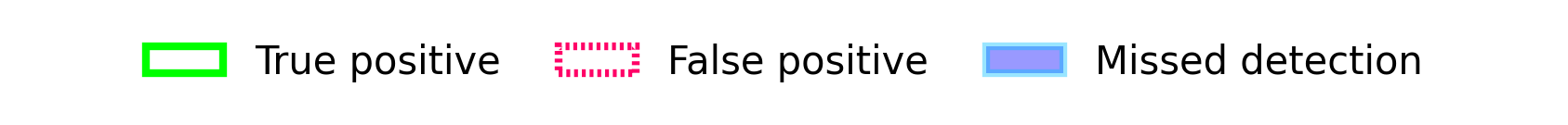} 
    \end{tabular}
    \vspace{0.0ex}  % spacing 
    \caption{Visualization of ablation studies. 
\textbf{Top row:} Comparison between~\DCP~and Separate Pooling, highlighting the benefit of unified object-background feature extraction. 
\textbf{Middle row:} Impact of different fusion strategies in the~\FusionHead—Cross-Attention, Addition, and Multiplication. 
\textbf{Bottom row:} Comparison between~\ProposalType~and standard Sparse Proposals, demonstrating the effect of incorporating background context.
}
\label{fig:Exp_ablation_visual_comparison}
\end{figure*}}

% The proposed DCP outperforms the separate pooling approach in terms of accuracy and inference speed, measured in frames per second (FPS).
% Thanks to its unified pooling strategy, both the object and background features are extracted from the same level of FPN feature maps, providing more consistent and aligned feature representations. 
% This consistency allows the cross-attention module in the~\FusionHead~to more effectively integrate object and background information, further leading to improved performance.
\colC{
The proposed DCP demonstrates clear improvements in both accuracy and inference speed (FPS) over the separate pooling baseline.
}
\colA{
\Figure~\ref{fig:Exp_ablation_visual_comparison}(a) and~\ref{fig:Exp_ablation_visual_comparison}(b) further confirm the above results, as the model incorporating DCP excellently detects all ship instances without false positives or missed detections.
}
\colC{
By jointly extracting object and background features from the same level of the FPN, it maintains consistent feature alignment. 
This consistency allows the~\FusionHead~to more effectively combine contextual information, leading to enhanced detection performance.
}
\colC{
In contrast, the separate pooling strategy processes object and background features independently, often resulting in features drawn from different FPN layers. 
Since background boxes tend to be larger, their features are typically pooled from shallower layers in the FPN than those of object boxes, leading to inconsistent feature levels. 
This misalignment weakens feature fusion consistency and can impair the~\FusionHead's ability to reason over object–background relations.
Additionally, by removing the need for two separate pooling operations, DCP improves inference speed, as confirmed in the experiments.}

\colA{
The performance of DCP is sensitive to the extension factor $\alpha$, a hyperparameter that scales the object's width and height to include surrounding background information, as detailed in Section~\ref{SubSection:Dual-Context-Pooling}. We found that setting $\alpha = 1.86 \approx 13/7$ yields better accuracy across the defined metrics, while preserving efficient training and inference times, as presented in~\Table~\ref{table:Exp_Extension_Factor}. 
Based on these findings, we adopt $\alpha = 1.86$ as the default in the model.
\begin{table}[htbp]
\centering
\caption{Effect of extension factor $\alpha$}
\label{table:Exp_Extension_Factor}
\renewcommand{\arraystretch}{1.2} % Adjust the value to increase the spacing
\setlength{\tabcolsep}{2.0pt}
\centering
\begin{tabular}{>
{\centering\arraybackslash}m{0.8cm}>
{\centering\arraybackslash}m{1.2cm}>
{\centering\arraybackslash}m{1.2cm}>
{\centering\arraybackslash}m{1.2cm}>
{\centering\arraybackslash}m{1.6cm}>
{\centering\arraybackslash}m{1.2cm}}
        
	\hline
	
        \colA{$\alpha$}& 
	\colA{AP}& 
	\colA{AP\textsubscript{50}}& 
        \colA{AP\textsubscript{75}}& 
	\colA{Training time (h)}&
	\colA{FPS} \\\hline
	
	\colA{9/7} & \colA{0.497} & \colA{0.898} & \colA{0.515} & \colA{4.18} & \colA{13} \\ 
	\colA{11/7} & \colA{0.499} & \colA{0.922} & \colA{0.498} & \colA{4.41} & \colA{13}\\ 
	\colA{13/7} & \colA{0.511} & \colA{0.927} & \colA{0.519} & \colA{4.47} & \colA{13}\\
	\colA{15/7} & \colA{0.489} & \colA{0.890} & \colA{0.487} & \colA{4.49} & \colA{13}\\
        \colA{17/7} & \colA{0.496} & \colA{0.912} & \colA{0.505} & \colA{4.54} & \colA{12}\\\hline

\end{tabular}

\raggedright
\end{table}
}

\subsection{Fusion Head Performance}

The~\FusionHead~combines features \colC{resulting from} the~\BackgroundHead~\colC{and the ~\ObjectHead~and feeds them to the regression layer.
We use a cross-attention mechanism for fusion (\Equation~\ref{eqn:FusionHead}) and validate it against standard methods such as element-wise addition and multiplication, as shown in~\Table~\ref{table:Exp_Fusion_Heads}.}

\begin{table}[htbp]
\centering
\caption{Experiment on Different Fusion Strategies for Fusion Head}
\label{table:Exp_Fusion_Heads}
\renewcommand{\arraystretch}{1.2} % Adjust the value to increase the spacing
\setlength{\tabcolsep}{2.0pt}
\centering
\begin{tabular}{>{\centering\arraybackslash}m{2.0cm}>{\centering\arraybackslash}m{1.8cm}>
{\centering\arraybackslash}m{1.8cm}>{\centering\arraybackslash}m{2.0cm}>{\centering\arraybackslash}m{2.3cm}}
	\hline
	Fusion type& 
	% AP\textsubscript{50}& 
    AP\textsubscript{50}Inshore& 
	Model size (M)& 
	Speed (millisecond)\\\hline
	 
	Cross-attention & 0.793 & 185.25 & 77.945\\ 
	Addition & 0.717 & 184.05 & 76.513 \\ 
	Multiplication & 0.705 & 184.05 & 76.513\\\hline
\end{tabular}

\raggedright
\end{table}

The proposed~\FusionHead~with cross-attention improves inshore $AP_{50}$ by 7.6\% over other fusion strategies, demonstrating its effectiveness in complex scenes.
\colA{
This performance gain is also evident in~\Figure~\ref{fig:Exp_ablation_visual_comparison}(c)–(e), where the cross-attention-powered~\FusionHead~successfully detects all ship instances, unlike addition and multiplication, which are prone to missed detections.}
Element-wise addition and multiplication fuse features in a fixed, position-wise manner, treating all spatial locations equally and lacking the ability to prioritize informative regions.
In contrast, cross-attention dynamically weighs feature importance and selectively integrates background cues based on their relevance to each ship target. 
This targeted fusion enhances object features with supportive contextual information—such as surrounding textures or spatial patterns—which reinforces object evidence and improves semantic consistency.
These enriched features lead to higher classification confidence for true positives, making them more likely to pass detection thresholds and reducing missed detections.
Despite its benefits, the cross-attention module introduces only 1.1M additional parameters and a 1.4ms increase in inference time, maintaining a favorable balance between performance and efficiency.

\subsection{Effectiveness of Background-Aware Proposals}
Background interaction is the key distinction of~\ModelName~compared to the baseline model,~\SparseRCNN. This interaction improves the contextual awareness of proposals, allowing more precise predictions using local background information.

\colC{To demonstrate the effectiveness of~\ProposalType~(\ProposalTypeAbbr), we compare the model's performance against an extended version of~\SparseRCNN, modified to support rotated object detection.}
Specifically, we extend the proposal boxes in~\SparseRCNN~\colC{to include an orientation parameter} and update its regression layer to predict offsets for those boxes.
The modifications are identical to those applied to the proposal box and regression layer in~\ModelName, as detailed in~\Section~\ref{SubSection:Rotated-Learnable-Proposals}~and~\Section~\ref{Subsubsection:Classfication-Regression-Module}.
The overall architectures of \colA{the proposed~\ModelName~}and the modified~\SparseRCNN~are closely similar, except \colC{that}~\ModelName~incorporates background features into proposals and employs a~\FusionHead~ whereas ~\SparseRCNN~ does not.
Hence, the modified~\SparseRCNN~serves as a strong baseline to emphasize the impact of excluding the background-aware mechanism in \colA{the proposed model}.

\begin{table}[htbp]
\centering
\caption{Effectiveness of background-aware proposals}
\label{table:Exp_Background_Aware}
\renewcommand{\arraystretch}{1.2} % Adjust the value to increase the spacing
\setlength{\tabcolsep}{2.0pt}
\begin{tabular}{>{\centering\arraybackslash}m{2.5cm}>{\centering\arraybackslash}m{1.2cm}>{\centering\arraybackslash}m{2.5cm}>{\centering\arraybackslash}m{1.7cm}}
\hline
Model& 
AP\textsubscript{50}& 
AP\textsubscript{50}Inshore& 
FPS \\\hline

\ModelName & 0.927\textbf{(+2.2\%)} & 0.793\textbf{(+6.4\%}) & 13 \\
\colB{Sparse R-CNN} & 0.905 & 0.729 & 25\\\hline

\end{tabular}
\end{table}

As shown in~\Table~\ref{table:Exp_Background_Aware}, it is clear that the proposed model benefits from background-background interaction, marking the effectiveness of~\ProposalTypeAbbr. 
Notably, in challenging inshore regions,~\ModelName~outperforms its counterpart with a 6.4\% higher $AP_{50}$.
The lower AP rates of the modified~\SparseRCNN~confirm that object-centric proposals struggle with ambiguity arising from the lack of contextual background information.
In contrast,~\ModelName, explicitly treats the background as a distinctive feature set and dynamically interacts it with proposal features. 
This interaction allows \colA{the proposed model} to adaptively fuse background and proposal features, enhancing contextual understanding and improving detection performance in complex environments.
\colA{The performance gap is further visualized in~\Figure~\ref{fig:Exp_ablation_visual_comparison}(f) and ~\Figure~\ref{fig:Exp_ablation_visual_comparison}(g), where the model employing~\ProposalType~is noticeably better at avoiding false positives due to its background-aware design.}

\subsection{Comparison to SOTA Detectors}

To verify the advanced performance of~\ModelName, we compare its performance against state-of-the-art (SOTA) one-stage, two-stage, and anchor-free detectors. 
These detectors were built on top of the MMRotate~\cite{MMRotate}~codebase.
Performance comparisons on the SSDD and RSDD-SAR datasets are shown in~\Table~\ref{table:performance_comparison_SSDD} and~\Table~\ref{table:performance_comparison_RSDD}, respectively

\begin{table*}[t]
\centering
\caption{Performance comparison on SSDD Dataset.}
\label{table:performance_comparison_SSDD}
\renewcommand{\arraystretch}{1.2} % Adjust the value to increase the spacing
\begin{tabular}{
		>{\centering\arraybackslash}m{1.0cm}
		>{\centering\arraybackslash}m{2.8cm}
		>{\centering\arraybackslash}m{1.0cm}
		>{\centering\arraybackslash}m{1.0cm}
		>{\centering\arraybackslash}m{1.0cm}
		>{\centering\arraybackslash}m{1.0cm}
		>{\centering\arraybackslash}m{1.0cm}
		>{\centering\arraybackslash}m{1.0cm}
	}
	\\
	\hline
	Type & 
	Model & 
	AP &
	AP\textsubscript{50} & 
	AP\textsubscript{75} &
	AP\textsubscript{50} Inshore &
	AP\textsubscript{50} Offshore &
	FPS
	\\\hline
	
	\multirow{4}{*}
	{Two-stage} 
	& R-Faster R-CNN & 0.503 & 0.903 & 0.506 & 0.789 & 0.905 & 40\\
	& RoI Transformer & 0.504 & 0.901 & \textbf{0.527} & 0.781 & 0.910 & 32\\
	& Gliding Vertex & 0.508 & 0.903 & 0.506 & 0.782 & 0.905 & 38\\ 
	& Oriented R-CNN & 0.509 & 0.899 & 0.513 & \colB{\textbf{0.794}} & 0.907 & 38\\ 
	\hline
	
	\multirow{4}{*}{One-stage} 
	& R-RetinaNet & \colD{\textbf{0.511}} & 0.893 & 0.515 & 0.731 & 0.908 & 48 \\
	& S2ANet & 0.501 & 0.883 & 0.513 & 0.763 & 0.908 & 36\\ 
	& R3Det & 0.475 & 0.881 & 0.481 & 0.678 & 0.891 & 36\\
	\hline
	
	\multirow{3}{*}{Anchor-free} 
	& CFA & 0.445 & 0.876 & 0.472 & 0.665 & 0.882 & 43\\
	& R-FCOS & 0.493 & 0.881 & 0.496 & 0.726 & 0.908 & \textbf{49} \\
	& Oriented RepPoints & 0.467 & 0.879 & 0.485 & 0.672 & 0.892 & 44 \\
	\hline
	
	% Proposed
	\multirow{1}{*}{} & \ModelName & \colD{\textbf{0.511}} & \textbf{0.927} & 0.518 & 0.793 & \textbf{0.972} & 16 \\
	\hline
	
	% Footnote
	\multicolumn{8}{l}{\scriptsize{Bold values indicate the highest performance for each metric.}} \\ \\

\end{tabular}

\end{table*}

\begin{table*}[t]
\centering
\caption{Performance comparison on RSDD-SAR Dataset.}
\label{table:performance_comparison_RSDD}
\renewcommand{\arraystretch}{1.2} % Adjust the value to increase the spacing
\begin{tabular}{
		>{\centering\arraybackslash}m{1.0cm}
		>{\centering\arraybackslash}m{2.8cm}
		>{\centering\arraybackslash}m{1.0cm}
		>{\centering\arraybackslash}m{1.0cm}
		>{\centering\arraybackslash}m{1.0cm}
		>{\centering\arraybackslash}m{1.0cm}
		>{\centering\arraybackslash}m{1.0cm}
		>{\centering\arraybackslash}m{0.5cm}
	}
	\\
	\hline
	Type & 
	Model & 
	AP &
	AP\textsubscript{50} & 
	AP\textsubscript{75} &
	AP\textsubscript{50} Inshore &
	AP\textsubscript{50} Offshore 
	\\\hline
	
	\multirow{4}{*}
	{Two-stage} 
	& R-Faster R-CNN & 0.473 & 0.886 & 0.450 & 0.644 & 0.898  \\
	& RoI Transformer & 0.501 & 0.900 & \textbf{0.539} & 0.664 & 0.909 \\
	& Gliding Vertex & 0.450 & 0.896 & 0.516 & 0.578 & 0.905 \\ 
	& Oriented R-CNN & 0.503 & 0.898 & 0.513 & 0.633 & 0.905 \\ 
	\hline
	
	\multirow{4}{*}{One-stage} 
	& R-RetinaNet & 0.478 & 0.877 & 0.423 & 0.599 & 0.891 \\
	& S2ANet & 0.502 & 0.898 & 0.477 & 0.592 & 0.892 \\ 
	& R3Det & 0.489 & 0.865 & 0.480 & 0.568 & 0.904 \\
	\hline
	
	\multirow{3}{*}{Anchor-free} 
	& CFA & 0.498 & 0.878 & 0.491 & 0.549 & 0.886\\
	& R-FCOS & 0.499 & 0.888 & 0.505 & 0.616 & 0.903\\
	& Oriented RepPoints & 0.492 & 0.879 & 0.496 & 0.578 & 0.899\\
	\hline
	
	% Proposed
	\multirow{1}{*}{} & \ModelName & \textbf{0.511} & \textbf{0.914} & 0.522 & \textbf{0.668} & \textbf{0.961}\\
	\hline
	
		% Footnote
	\multicolumn{7}{l}{\scriptsize{Bold values indicate the highest performance for each metric.}} \\
	
\end{tabular}
\end{table*}

%The proposed~\ModelName~demonstrates superior performance over other detectors across most metrics and various ship detection scenarios, including both inshore and offshore environments, for both datasets.
%In SSDD test set,~\ModelName

On the SSDD dataset,~\ModelName~demonstrates consistently strong performance across defined evaluation metrics.
\colD{Specifically,~\ModelName~shares the highest overall AP of 0.511 with R-RetinaNet, outperforming all other detectors across mixed inshore and offshore scenes.
In offshore regions,~\ModelName~achieves an exceptional $AP_{50}$ of 0.972, surpassing the closest competitor, RoI Transformer, by 6.2\%.
}
\colB{
In inshore regions,~\ModelName~achieves a competitive $AP_{50}$ of 0.793, with only a 0.1\% gap to the best-performing method, Oriented R-CNN.}
Similarly, on the RSDD-SAR dataset,~\ModelName~maintains its leading position with the highest $AP$ of 0.511 and excels in both offshore ($AP_{50}$ = 0.961) and inshore ($AP_{50}$ = 0.668) scenarios, highlighting its adaptability to diverse environmental conditions.
This strong performance on both SSDD and RSDD-SAR datasets highlights~\ModelName's~capability to handle datasets with varying characteristics, achieving consistent performance across different scenarios.

The experimental results on both datasets confirm that~\ModelName~is effective in both inshore and offshore scenarios, showcasing its ability to adapt to varying environmental complexities. 
Offshore regions, often characterized by sparse backgrounds, require precise object detection with minimal contextual interference, where~\ModelName~performs exceptionally well. 
Inshore regions, on the other hand, present challenges such as cluttered backgrounds and higher object densities. 
Despite these challenges,~\ModelName~demonstrates strong detection capabilities, outperforming other methods in most metrics. 
Furthermore,|\ModelName~achieves the second-highest $AP_{75}$ on both SSDD and RSDD-SAR datasets, demonstrating its strong capability in precise object localization under stricter IoU thresholds.
Although its inference speed (16 FPS) is slower compared to other two-stage models like R-Faster R-CNN (40 FPS), this trade-off is justified by the substantial improvements in detection accuracy across all metrics.

\begin{figure*}[!htbp]
\includegraphics[width=0.95\textwidth,center, trim={0 0 0 0}]{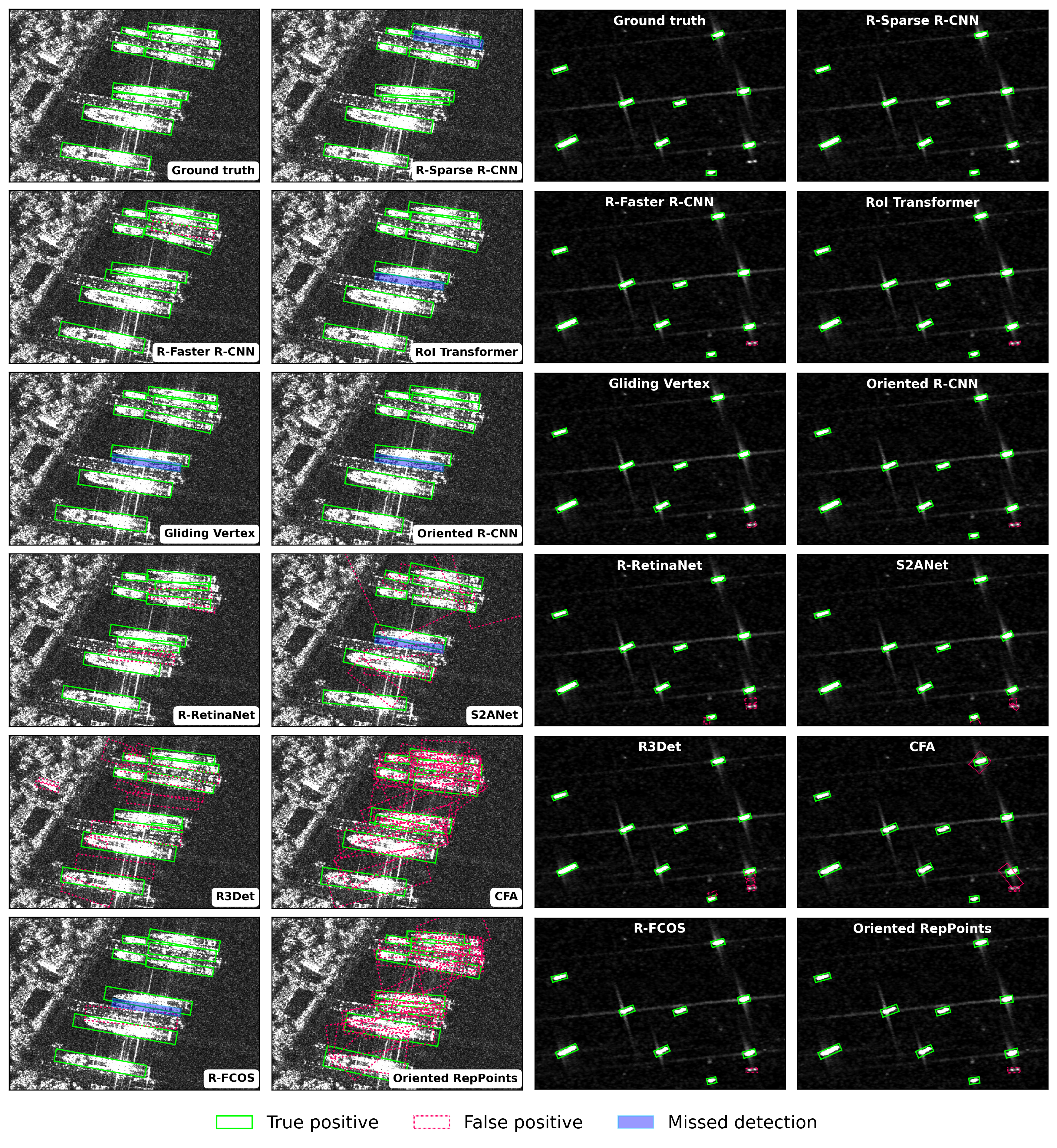}\caption
{Comparison of detection results on SSDD dataset.}
\label{fig:Visual_Comparison_SSDD}
\end{figure*}

\begin{figure*}[!htbp]
\includegraphics[width=0.8\textwidth,center, trim={0 0 0 0}]{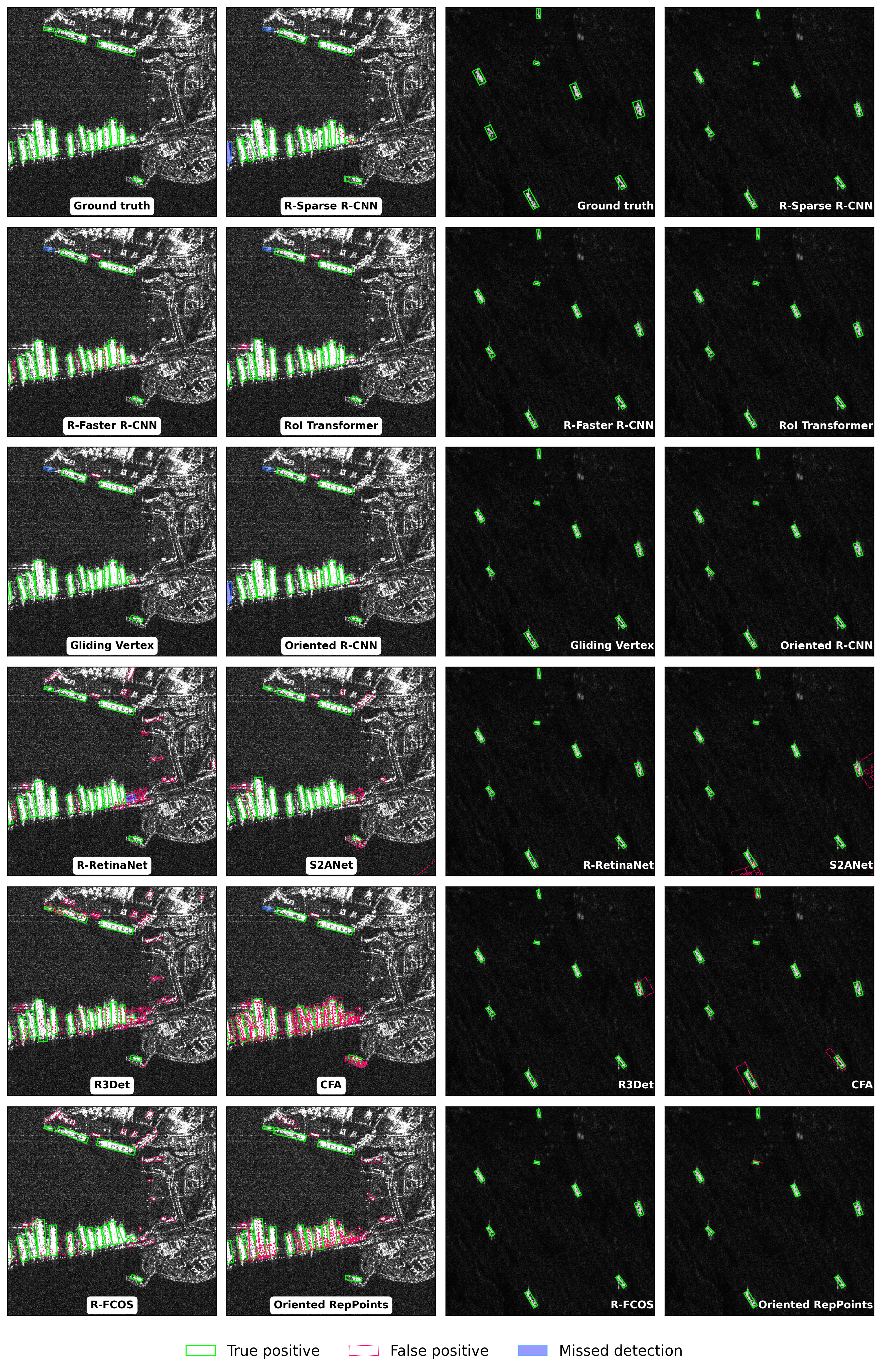}
\caption
{Comparison of detection results on RSDD-SAR dataset.}
\label{fig:Visual_Comparison_RSDD}
\end{figure*}

Finally, visual comparisons of test results on the SSDD and RSDD-SAR datasets are presented in~\Figure~\ref{fig:Visual_Comparison_SSDD}~and~\Figure~\ref{fig:Visual_Comparison_RSDD}, respectively. 
These visual comparisons further confirm the accuracy metrics, as~\ModelName~demonstrates a closer match to the ground truth boxes on both inshore and offshore scenes.
Furthermore, the performance improvement is more evident in inshore regions, where R-Sparse R-CNN effectively reduces false detections.
% This highlights the advantage of incorporating background-background interactions into our model, which enhances its robustness and accuracy in challenging detection tasks.
\colC{
This reduction in false positives reflects the advantage of employing sparse proposal representations in SAR images containing ships, where targets are typically sparse in space. 
Sparse proposals help reduce false positives by limiting the number of candidate regions to a small, high-quality set, making the model less susceptible to false detections.
}

% This alignment between quantitative metrics and qualitative results highlights the robustness and reliability of the proposed model.

In summary, the experimental results highlight the effectiveness of the proposed~\ModelName~for oriented ship detection in SAR images.
By eliminating the need for RPN training, dense anchor design, and NMS post-processing, this approach simplifies the detection pipeline while maintaining competitive performance. 
These findings validate the feasibility of a streamlined model architecture for more efficient and accurate SAR ship detection.
Furthermore, \colC{experimental results} validate the advantage of leveraging background information to enhance detection accuracy through~\ProposalType~mechanism.
This approach enables the model to dynamically learn the relationship between objects and their backgrounds, promoting a richer contextual understanding within the model.
Additionally, the proposed~\DCP~presents an efficient method for unified pooling of background and object features, enabling faster execution and enhanced model accuracy.
This approach is adaptable for use in other implementations requiring object-background pooling, making it a significant contribution to oriented SAR ship detection field.

\subsection{Performance on Large-Scale Image}
\label{Performance_on_large_scale_image}

\begin{figure*}[!htbp]
\includegraphics[width=0.90\textwidth,center, trim={0 0 0 0}]{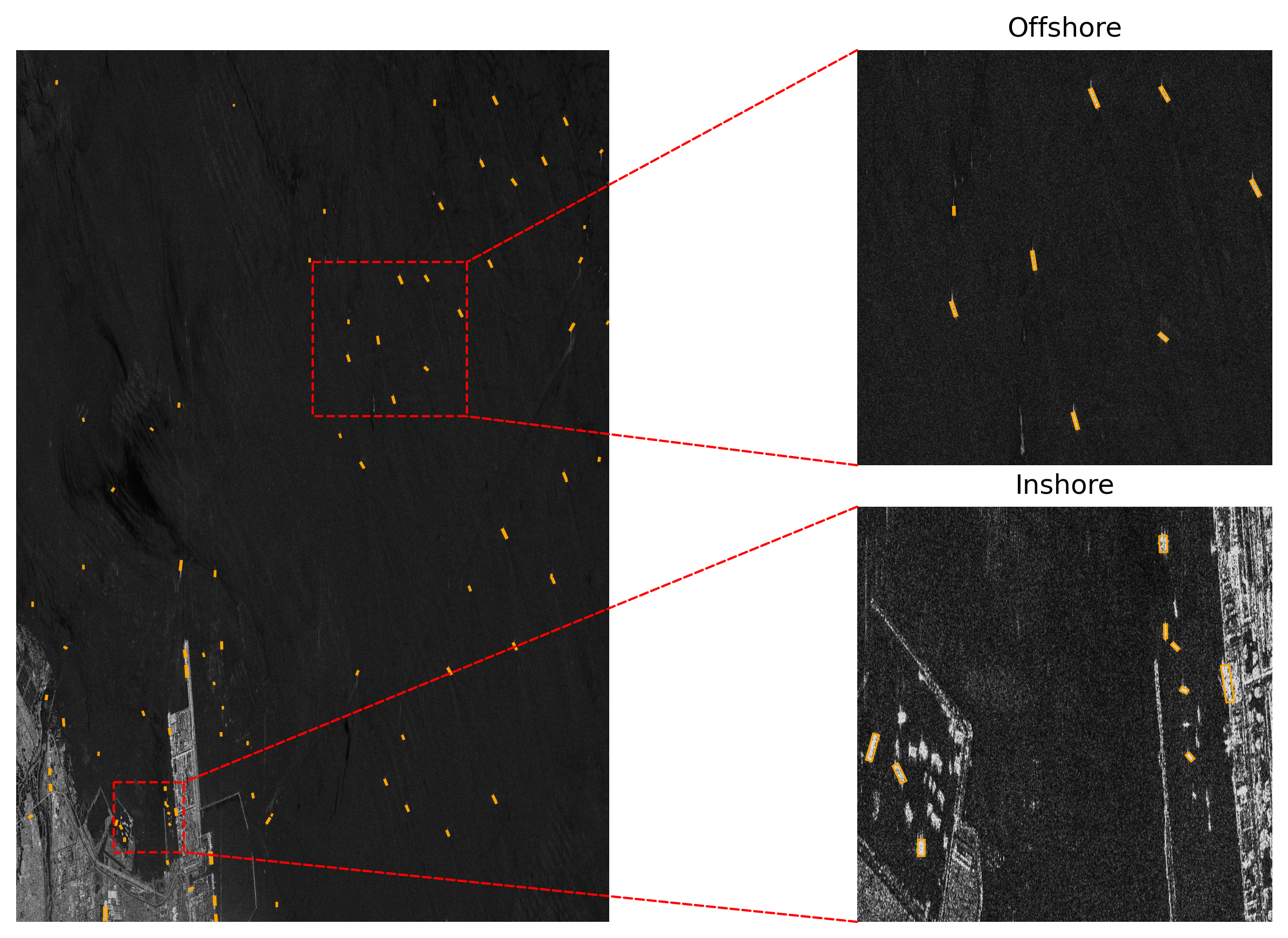}
\caption
{\colA{Model performance on large-scale Gaofen-3 (GF-3).}}
\label{fig:Large_scale_performance}
\end{figure*}

\colA{
To further assess the robustness of the proposed method, we conduct an additional experiment on a large-scale GaoFen-3 (GF-3) SAR image from the RSDD-SAR validation set.
The image spans both inshore and offshore regions, with a resolution of 3 meters and dimensions of $8,500 \times 12,500$ pixels. 
We directly apply the model trained on the SSDD dataset to this image. 
As shown in~\Figure~\ref{fig:Large_scale_performance}, the model successfully detects the majority of ship targets in the offshore region, suggesting strong generalization capability in open-sea scenarios, even under domain shift.
}

\colA{
In the inshore region, the model occasionally misses small vessels, which is a known limitation of modern detectors when dealing with increased scene complexity and domain shift. 
This performance drop can be attributed to two main factors: (1) the presence of artificial coastal structures that increase background clutter, and (2) the limited representation of inshore scenarios in the SSDD training set. Despite these limitations, the model demonstrates robust detection in offshore environments and highlights potential directions for improving performance in more challenging inshore scenarios.
}

\section{Conclusion}
\label{Section:Conclusion}
\colA{
We proposed R-Sparse R-CNN, a novel framework for oriented ship detection in SAR imagery that leverages sparse learnable proposals enriched with background context, referred to as~\ProposalType~(\ProposalTypeAbbr).
The adoption of sparse learnable proposals concept ensures a streamlined design for the pipeline eliminating the need for dense anchors, RPN training, as well as NMS post-processing.
Additionally, the incorporation of background contextual information on BAP improves the} \colC{model's detection accuracy} \colA{in both inshore and offshore regions, as it enables the model to dynamically learn object–background relationships and promotes richer contextual understanding.}

\colA{
To complement~\ProposalType, we introduced~\DCP~(DCP), a unified pooling strategy that efficiently extracts both object and background features in a single operation. Unlike separate pooling schemes, DCP avoids redundant computation and ensures both features are extracted from the same Feature Pyramid Network (FPN) layer of feature maps. This design provides better alignment and consistent representation between object and background features regardless of proposal sizes, which evidently improves detection accuracy. }

\colA{
At the core of object–background learning in~\ModelName, we design an~\InteractionModule~consisting of two~\InteractionHeads~and a~\FusionHead. The~\InteractionHeads~first model object–object and background–background interactions between Regions of Interest (RoIs) and proposals. 
The~\FusionHead~then combines the resulting features through cross-attention-based reasoning, enabling more informed proposal refinement and ultimately improving performance.
}

\colA{
Extensive experiments on SSDD and RSDD-SAR datasets validate the robustness of~\ModelName~in detecting SAR ships with arbitrary orientations across both inshore and offshore regions. 
We anticipate that~\ModelName~can be extended to other domains of oriented object detection, offering significant contributions to the field of remote sensing.
}

\section*{Acknowledgements}
    
Kamirul Kamirul is fully funded by Lembaga Pengelola Dana Pendidikan (LPDP), Ministry of Finance of Republic of Indonesia.
Odysseas A. Pappas and Alin M. Achim are funded by the Engineering and Physical Sciences Research Council - Impact Acceleration Account (EPSRC - IAA) University of Bristol 2022 under reference number EP/X525674/1.

\bibliographystyle{IEEEtran}
\bibliography{references}

\begin{IEEEbiography}[{\includegraphics[width=1in,height=1.25in,clip,keepaspectratio]{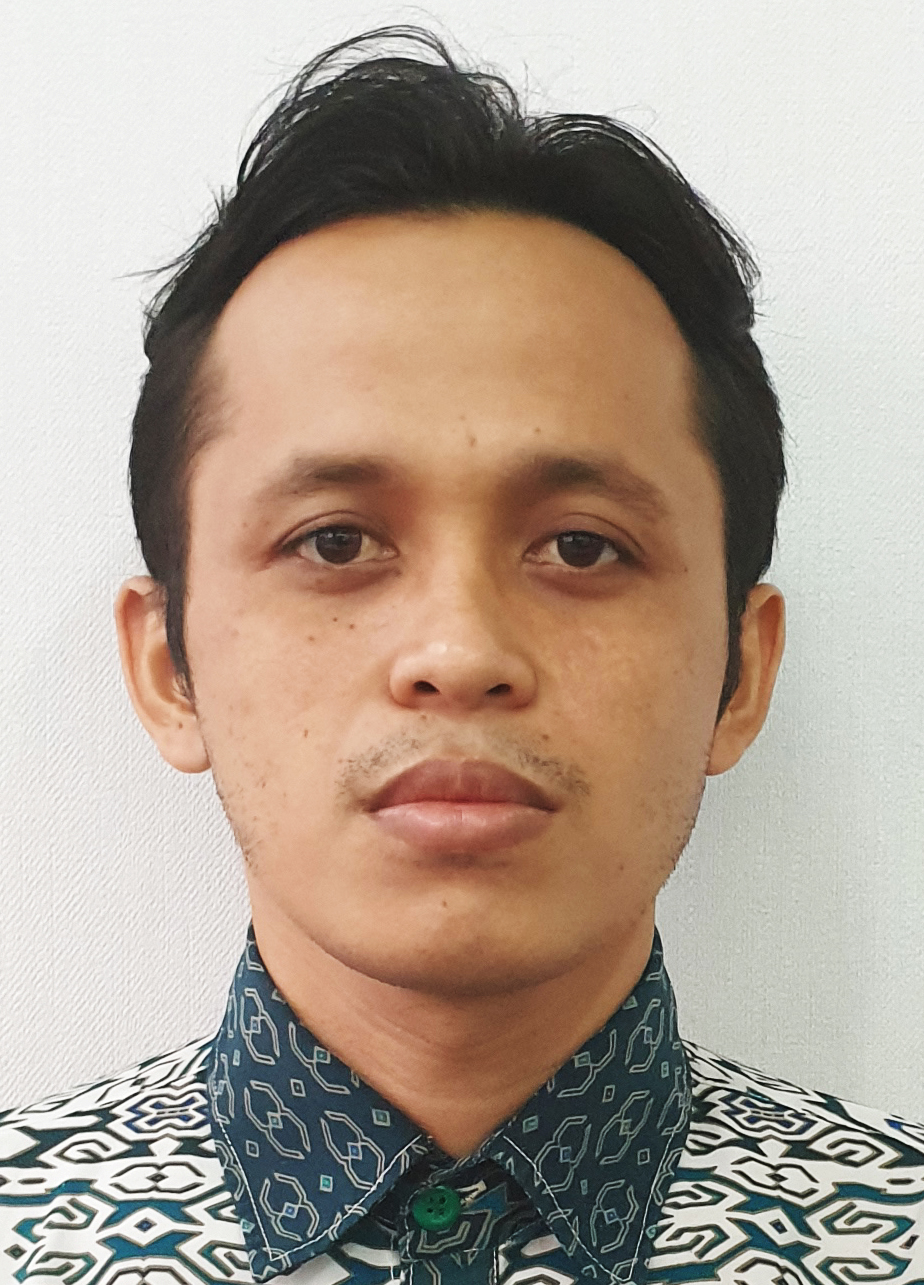}}]{Kamirul Kamirul} (Student Member, IEEE) received the B.Sc. degree in Physics from Universitas Tanjungpura, Pontianak, Indonesia, in 2014, and the M.Sc. degree in Physics from Institut Teknologi Bandung, Indonesia, in 2017. 
He is currently pursuing a Ph.D. at the Visual Information Laboratory, University of Bristol. 
Additionally, he serves as a Satellite Data Processing Engineer at the National Research and Innovation Agency of Indonesia (BRIN) since 2018. 
His research interests include ship detection and identification in SAR imagery.
\end{IEEEbiography}

\begin{IEEEbiography}[{\includegraphics[width=1in,height=1.25in,clip,keepaspectratio]{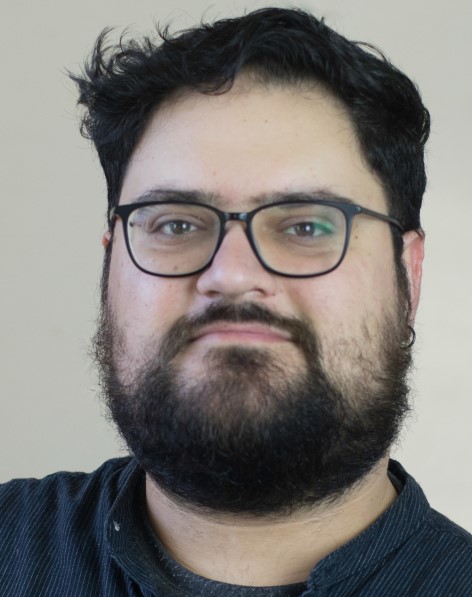}}]{Odysseas A. Pappas} received the M.Eng. degree in electrical and electronic engineering and the Ph.D. degree in communications and signal processing from the University of Bristol, Bristol, U.K., in 2012 and 2017, respectively.
He has been a Research Associate with the Visual Information Laboratory, University of Bristol, since 2018. 
In 2019, he joined the School of Earth Sciences, University of Bristol. His research interests focus on remote sensing and include image fusion, target detection, superpixel segmentation, statistical modeling, anomaly detection, and inverse problems in imaging.
\end{IEEEbiography}

\begin{IEEEbiography}[{\includegraphics[width=1in,height=1.25in,clip,keepaspectratio]{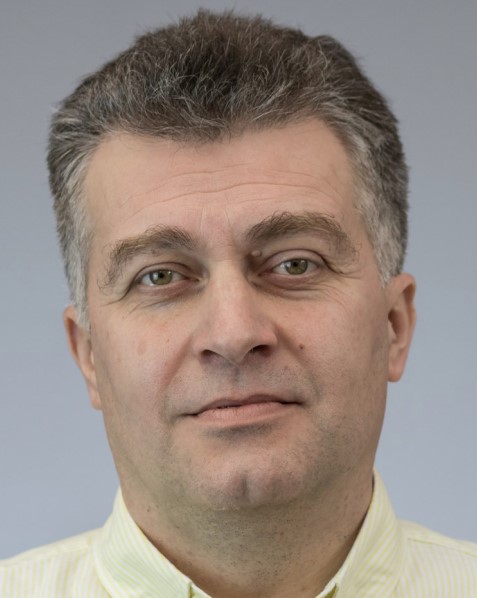}}]{Alin M. Achim}(Senior Member, IEEE) received the B.Sc. and M.Sc. degrees in electrical engineering from the “Politehnica” University of Bucharest, Bucharest, Romania, in 1995 and 1996, respectively, and the Ph.D. degree in biomedical engineering from the University of Patras, Patras, Greece, in 2003.
He received the European Research Consortium for Informatics and Mathematics (ERCIM) Post-doctoral Fellowship, which he spent with the Institute of Information Science and Technologies (ISTI-CNR), Pisa, Italy, and the French National Institute for Research in Computer Science and Control (INRIA) Sophia Antipolis, Biot, France. 
In October 2004, he joined the Department of Electrical and Electronic Engineering, University of Bristol, Bristol, U.K., as a Lecturer, where he became a Senior Lecturer (Associate Professor) in 2010 and a Reader in biomedical image computing in 2015. 
Since August 2018, he has been holding the Chair of Computational Imaging, University of Bristol. 
He has coauthored over 130 scientific publications, including more than 35 journal articles. 
His research interests include statistical signal, and image and video processings with a particular emphasis on the use of sparse distributions within the sparse domains and with applications in both biomedical imaging and remote sensing.
Dr. Achim is an elected member of the Bio Imaging and Signal Processing Technical Committee of the IEEE Signal Processing Society, an affiliated member (invited) of the IEEE Signal Processing Society’s Signal Processing Theory and Methods Technical Committee, and a member of the IEEE Geoscience and Remote Sensing Society’s Image Analysis and Data Fusion Technical Committee. 
He is a Senior Area Editor of the IEEE TRANSACTIONS ON IMAGE PROCESSING, an Associate Editor of the IEEE TRANSACTIONS ON COMPUTATIONAL IMAGING, and an Editorial Board Member of Remote Sensing (MDPI).
\end{IEEEbiography}

\vspace{11pt}

\end{document}